\documentclass[manuscript,screen,nonacm]{acmart}
\setcopyright{none}
\renewcommand\footnotetextcopyrightpermission[1]{}
\AtBeginDocument{%
  }

\usepackage{algorithmic}
\usepackage{algorithm}

\newcommand{\cov}{\mathrm{cov}}
\begin{document}

\title{Bayesian Nonparametric Dynamical Clustering of Time Series}

\author{Adri\'an P\'erez-Herrero}
\email{adrian.perez.herrero@usc.es}
\orcid{0000-0003-2085-1075}
\affiliation{%
  \institution{University of Santiago de Compostela}
  \city{Santiago de Compostela}
  \state{A Coruña}
  \country{Spain}
}

\author{Paulo F\'elix}
\orcid{0000-0002-1974-5665}
\affiliation{%
  \institution{University of Santiago de Compostela}
  \city{Santiago de Compostela}
  \state{A Coruña}
  \country{Spain}
}
\email{paulo.felix@usc.es}

\author{Carl Henrik Ek}
\orcid{0000-0003-1302-6309}
\affiliation{%
  \institution{University of Cambridge}
  \city{Cambridge}
  \country{United Kingdom}
}
\email{che29@cam.ac.uk}

\author{Jes\'us Presedo}
\orcid{0000-0001-8066-4760}
\affiliation{%
  \institution{University of Santiago de Compostela}
  \city{Santiago de Compostela}
  \state{A Coruña}
  \country{Spain}
}
\email{jesus.presedo@usc.es}


\begin{abstract}
  We present a method that models the evolution of an unbounded number of time series clusters by switching among an unknown number of regimes with linear dynamics. We develop a Bayesian non-parametric approach using a hierarchical Dirichlet process as a prior on the parameters of a Switching Linear Dynamical System and a Gaussian process prior to model the statistical variation in amplitude and temporal alignment within each cluster. By modelling the evolution of time series patterns, the method avoids unnecessary proliferation of clusters in a principled manner. We perform inference by formulating a variational objective for offline and online scenarios, enabling efficient learning through optimisation. We illustrate the versatility and effectiveness of the approach through applications to electrocardiogram (ECG) morphology clustering, respiration estimation from ECG, directional wave-spectra interpretation, and pedestrian trajectory identification.
\end{abstract}



\maketitle

\section{Introduction}
\label{sec:introduction}

Time series data analysis has come to pervade all scientific and technological domains, driven by the need to understand change over time. With the growing availability of such data, machine learning has assumed an increasingly central role in a wide variety of tasks which fall under the category of pattern recognition. Particularly, there is growing interest in identifying similar behaviours in time series data as a preliminary step towards generating insights into the dynamics of the underlying processes. Some recent methodologies can be found for characterising sea wave conditions~\citep{hamilton2010}, transcriptome-wide gene expression profiling~\citep{mcdowell2018}, selecting stocks with different share price performance~\citep{nieto2014}, and discovering human motion primitives~\citep{zhou2013}.

One of the main challenges in this setting is the dynamic nature of time series data, which often exhibits evolving trends, shifts over time and non-stationary patterns. While traditional clustering techniques effectively cope with sequence pattern analysis, they often focus on task-specific performance rather than the more fundamental question of describing the dynamics that generated the time series~\citep{rabiner_1989}. Recovering the generative mechanism enables us to learn more about the data source (the real-world process) and to devise methods for solving prediction, recognition or identification tasks.

Linear dynamical systems (LDS) have been successfully applied in describing dynamical phenomena by considering continuous hidden state variables with Markovian dynamics given by Gaussian linear relationships between consecutive states. Furthermore, when time series data exhibit nonlinear or even non-stationary dynamics, switching linear dynamical systems (SLDS) are proposed as a natural generalisation, where each state of a hidden Markov model (HMM) is associated with a linear dynamical process, so the discrete transition between states switches between different linear regimes~\citep{harrison1976,hamilton1989}. Applications of SLDS can be found in econometrics~\citep{kim1994}, human motion~\citep{pavlovic2000}, or target tracking~\citep{li2005}.

Model identification is a major issue in describing dynamics from a finite number of observations. Most existing methods for learning SLDS rely on either setting the number of different linear regimes~\citep{ghahramani_variational_2000}, or including a change-point detection procedure for the addition of a new regime~\citep{paoletti2007}. Fox {\em et al.}~\citep{fox_2011} relaxed this assumption to allow an unbounded number of regimes by proposing a Bayesian nonparametric approach. They make use of the hierarchical Dirichlet process (HDP) of Teh {\em et al.}~\citep{teh_2006} as a prior for the HMM. By introducing a prior over the addition of new regimes, Bayesian inference allows a natural trade-off between model complexity and data-fit in the posterior distribution, while leveraging previous knowledge of the dynamics.

A second challenge concerns the representation of the patterns themselves. Gaussian processes (GPs) provide some interesting properties for clustering tasks, such as a principled quantification of uncertainty, the ability to encode our previous beliefs about the function properties, such as smoothness, periodicity, or stationarity and a natural Bayesian interpretation~\citep{mackay1992,rasmussen_gaussian_2006}. This makes them well-suited to represent clusters of time-series segments as functional objects rather than fixed templates.

Nevertheless, while GPs capture amplitude variability, temporal variability falls outside their standard formulation. Temporal misalignment is commonly addressed using dynamic time warping (DTW)~\citep{sakoe1978} and related pairwise similarity methods. More recent GP-based alignment approaches either learn a nonlinear transformation of the GP output~\citep{snelson_2003,lazaro_2012}, or learn a monotonic function that warps the input space~\citep{duncker_2018,kazlauskaite_2019}. In the setting considered here, misalignment is not treated only as a nuisance caused by noisy sampling; it may be part of the generative mechanism itself and may carry application-specific information. This motivates modelling temporal alignment as a latent component of the model rather than as a preprocessing step.

In this paper, we propose a Bayesian nonparametric method for clustering time-series segments of different lengths, assuming that they are realisations of an underlying dynamic process. We consider the challenges above not as independent issues, but as interacting aspects of a single sequential learning problem: when a new segment is observed, the model must decide whether it belongs to a previously identified morphology or whether it represents a new one, while updating the corresponding morphological representation without drastically forgetting what that morphology meant before. This modelling task can be summarised by four inferential queries. First, given the last assignment to each cluster, what should a future instance of that cluster look like, and how uncertain is that prediction? This requires a functional predictive representation with calibrated uncertainty, which motivates the use of GP emissions. Second, given the new observation and the predictions from the existing clusters, should the segment be assigned to one of them, or should a new cluster be created? This requires a nonparametric prior over cluster identities, which motivates the HDP construction. Third, should the observation be aligned in order to better assess its agreement with the existing morphologies? This motivates a latent GP-based warping function, so that temporal alignment is performed as part of the generative model rather than as an external preprocessing step. Fourth, once the segment has been assigned to a cluster, how should the representation of that cluster be updated? This motivates the use of cluster-specific linear dynamical systems, which allow morphologies to evolve gradually while preserving their identity.

The remainder of the paper is organised as follows. Section~\ref{sec:notation} introduces the notation. Section~\ref{sec:problem} presents ECG heartbeat clustering as the motivating application. Section~\ref{sec:background} reviews the methodological background. Section~\ref{sec:generative_model} describes the HDP-GPC generative model, and Section~\ref{sec:variational_inference} presents the offline and online variational inference algorithms. Section~\ref{sec:hyperparameters} analyses the role of the main hyperparameters. Section~\ref{sec:applications} reports the experimental results, and Section~\ref{sec:related_work} discusses related formulations.

\section{Notation}
\label{sec:notation}
Consider a dataset $\mathcal{Y}=\{(\mathbf{t}_n,\mathbf{y}_n)\}_{n=1}^{N}$ of time-series segments, where $\mathbf{t}_n=(t_{ni})_{i=1}^q$ denotes an indexing time vector and $\mathbf{y}_n=(y_{ni})_{i=1}^q$ denotes a vector of real values. We are assuming that all the segments have the same number of samples to keep the notation uncluttered, although this is not a limitation of the present proposal. The result of the learning process is a generative model which captures the dynamics displayed by a number of evolving morphologies discovered within the dataset. These dynamics are summarised by a sequence of switch states $\{s_n\}_{n=1}^{N}$ evolving according to a discrete Markov transition structure, thus enabling us to effectively approximate the underlying dynamics through transitions between multiple linear regimes. Each linear regime makes evolve a Gaussian process $\{f_n\}_{n=1}^{N}$ representing each distinctive morphology. Before observation, these morphologies are assumed to be distorted according to another Gaussian process model $\{g_n\}_{n=1}^{N}$. 

\section{Motivating Example}
\label{sec:problem}

Before introducing the general model, we use electrocardiogram (ECG) heartbeat clustering as a motivating example to make the inferential requirements concrete. The ECG represents a recording of the electrical activity of the heart that is captured non-invasively from the surface of the body. An ECG signal comprises several significant waves and intervals, reflecting the anatomy and physiology of the conduction system of the heart. As such, it provides a rich source of information to analyse cardiovascular pathology, and even other pathologies that can affect electrocardiographic tracing.

ECG heartbeat clustering aims at dividing an ECG recording into a set of beat clusters, based on morphological similarity. Heartbeat clustering allows the identification of those beats that share a common activation origin and conduction pathway through the cardiac tissue, having proven to be an effective tool in diagnosing conduction disorders.

\begin{figure}
    \centering
    \includegraphics[width=\textwidth]{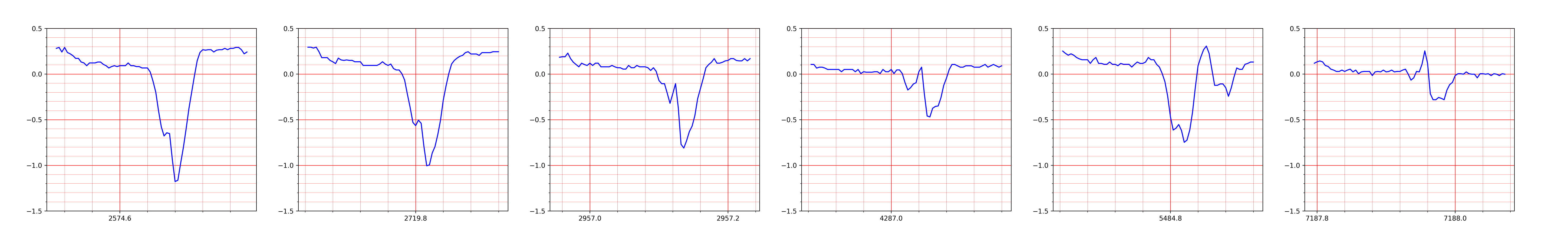}
    \caption{Several heartbeat examples from record e1302 of European ST-T Database~\citep{taddei_1992}. All of them were manually annotated as Normal except the fifth one, which was annotated as Ventricular~\citep{goldberger_2000}.}
    \label{fig:ecg_e1302}
    \Description[ECG examples with one distinct beat]{
    Line plot showing several ECG heartbeat segments from the same record. All beats except the fifth one have similar QRS-like morphology but with a considerable variation in amplitude, with a narrow positive peak followed by a negative deflection, while the fifth segment has a visibly different shape corresponding to a ventricular annotation.
    }
\end{figure}

Figure~\ref{fig:ecg_e1302} shows a great deal of variability among different realisations of the same conduction pattern through the cardiac tissue across time. There are two main sources of variability: on one hand, the stochastic nature of electrical propagation at the microscopic level along the cellular constituents of conduction pathways~\citep{madison1995}; and on the other hand, the available evidence on a measurable impact of different mechanical, neurological, metabolic and hormonal processes, among others, on the electrocardiographic tracing~\citep{malik_1996}. This example instantiates the four inferential questions introduced above. For each newly observed beat, the model must predict the expected morphology of previously identified beat families, decide whether the beat belongs to one of them or should instantiate a new family, compensate for the temporal misalignment of the QRS complex, and update the corresponding morphology without turning gradual physiological variability into redundant clusters.  We hypothesised that this variability results in an observable evolution of the heartbeat morphology that can be properly modelled by the present proposal. 

This example instantiates the four inferential questions introduced above. For each newly observed beat, the model must predict the expected morphology of the previously identified beat families, decide whether the beat belongs to one of them or should instantiate a new family, compensate for the temporal misalignment of the QRS complex, and update the corresponding morphology without turning gradual physiological variability into redundant clusters. We hypothesise that this variability produces an observable evolution of heartbeat morphology and that this evolution can be properly captured by the proposed Bayesian nonparametric dynamical model.

\section{Background}
\label{sec:background}
\subsection{Gaussian Processes}
A {\em Gaussian process} (GP) is defined as a collection of random variables, any finite number of which have a joint Gaussian distribution~\citep{rasmussen_gaussian_2006}. A GP can be indexed by some continuous set (e.g., time), which can be used to define a prior over an unknown function $f(t)$. A GP is fully specified by its mean function $m(t)$ and covariance function $k(t,t')$, and we will write $f(t) \sim \mathcal{GP}(m(t),k(t,t^{\prime}))$. Choosing the covariance function is a sort of model selection process that relies upon the objective of the modelling and the characteristics of the time series: by selecting a smooth covariance function, we can model smooth functions, by selecting a periodic covariance function, we can model periodic functions, etc. Ultimately, the covariance function determines almost all the properties of the resulting GP. We denote by $k_{\theta}(t,t^{\prime})$ the particular choice of the covariance function given by specific parameters $\theta$.

GPs are commonly used in regression tasks, consisting of learning from a dataset with data pairs $(t_i,y_i)_{i=1}^q$ where $\mathbf{t}=(t_1,...,t_q)$ denotes an input vector and $\mathbf{y}=(y_1,...,y_q)$ denotes a target vector of real values. We assume that each $y_i$ is a noisy version of the hidden $f(t_i)$, so that $\mathbf{y}=f(\mathbf{t})+\varepsilon$, with $\varepsilon \sim \mathcal{N}(0,\sigma^2)$, assuming additive identically distributed Gaussian noise. We denote by $f_i \triangleq f(t_i)$ the random variable representing the value of the function $f(t)$ at the input $t_i$. By the definition of GP, any finite number of random variables  $\mathbf{f}$ is distributed as $\mathbf{f}|\mathbf{t} \sim \mathcal{N}\left(\mathbf{f}|m(\mathbf{t}), K^{\theta}_{\mathbf{t},\mathbf{t}}\right)$, where $K^{\theta}_{\mathbf{t},\mathbf{t}}$ is a $q\times q$ matrix of components $k_{\theta}(t_i,t_j)$. Thus, prediction on new test points $\mathbf{t}^*$ can be computed from the joint distribution
\begin{equation}
 \left[\begin{array}{c} \mathbf{y}\\ \mathbf{f}^* \end{array}\right] \sim \mathcal{N}\left( \mathbf{0},\left[ \begin{array}{cc} K^{\theta}_{\mathbf{t},\mathbf{t}}+\sigma^2 I & K^{\theta}_{\mathbf{t},\mathbf{t}^*}\\ K^{\theta}_{\mathbf{t}^*,\mathbf{t}} & K^{\theta}_{\mathbf{t}^*,\mathbf{t}^*} \end{array}\right] \right).
\label{eq:gaussian_process}
\end{equation}
Deriving the conditional distribution, the predictive equations for GP regression are given by 
\begin{equation}
    \mathbf{f}^* | \mathbf{t}^*, \mathbf{t},  \mathbf{y} \sim \mathcal{N}(\mathbb{E}[\mathbf{f}^*], \text{cov}\left( \mathbf{f}^*\right)),
\end{equation}
where the predictive mean and covariance are
\begin{equation}
\begin{array}{rl}
    \mathbb{E}[\mathbf{f}^*] & = K^{\theta}_{\mathbf{t}^*,\mathbf{t}} \left[K^{\theta}_{\mathbf{t},\mathbf{t}}+\sigma^2 I\right]^{-1}\mathbf{y} \\
    \text{cov}( \mathbf{f}^*) & = K^{\theta}_{\mathbf{t}^*,\mathbf{t}^*} -K^{\theta}_{\mathbf{t}^*,\mathbf{t}} \left[K^{\theta}_{\mathbf{t},\mathbf{t}}+\sigma^2 I\right]^{-1}K^{\theta}_{\mathbf{t},\mathbf{t}^*}
\end{array}
\end{equation}
The computational complexity is dominated by the matrix inversion, which scales as $\mathcal{O}(q^3)$. To mitigate this, the information contained in the original $q$ input points can be projected onto a smaller set of $p<q$ inducing points (not necessarily a subset of the former), resulting in $\mathbf{f}_n^p\sim \mathcal{N}(m(\mathbf{t}^p),K^{\theta}_{\mathbf{t}^p,\mathbf{t}^p})$ latent variables, and thus benefiting computing efficiency. The linear Gaussian relation 
\begin{equation}
\mathbf{f}_n|\mathbf{f}_n^p \sim \mathcal{N}(\mathbf{f}_n|K^{\theta}_{\mathbf{t},\mathbf{t}^p}K^{\theta ~~-1}_{\mathbf{t}^{p},\mathbf{t}^p}\mathbf{f}_n^p,
K^{\theta}_{\mathbf{t},\mathbf{t}}-K^{\theta}_{\mathbf{t},\mathbf{t}^p}K^{\theta~~ -1}_{\mathbf{t}^p,\mathbf{t}^p}K^{\theta}_{\mathbf{t}^p,\mathbf{t}})
\label{eq:seeger_approach}
\end{equation}
where $\mathbb{E}[\mathbf{f}_n|\mathbf{f}^p_n]=K^{\theta}_{\mathbf{t},\mathbf{t}^p}K^{\theta ~~-1}_{\mathbf{t}^{p},\mathbf{t}^p}\mathbf{f}_n^p$, can be proven to be the KL-optimal projection, i.e., the approximating distribution $q(\mathbf{y}_n|\mathbf{f}_n^p)=\mathcal{N}(\mathbf{y}_n|\mathbb{E}[\mathbf{f}_n|\mathbf{f}^p_n],\sigma^2 I)$ allows us to minimize $\mathrm{KL}(q(\mathbf{f}_n|\mathbf{y}_n)||p(\mathbf{f}_n|\mathbf{y}_n))$ ~\citep{Seeger2005phd}. This approach also allows us to make predictions on new, different test points $\mathbf{t}^*$ from the joint distribution of $\mathbf{f}_n$ and the latent variables at the test locations $\mathbf{f}_n^*$ under the prior, by integrating over $\mathbf{f}_n^p$.

\subsection{GP-based temporal alignment}
Following Kazlauskaite {\em et al.}~\citep{kazlauskaite_2019}, we consider each sequence to be generated from a latent function $f_n(\mathbf{t}_n)$ and a monotonic warping $g_n(\mathbf{t}^0_n)$ as $\mathbf{y}_{n}=f_n(g_n(\mathbf{t}^0_{n}))+\varepsilon_{n}$ where $\varepsilon_n \sim \mathcal{N}(0,\sigma_{n}^2)$. The latent function $f_n$ can be drawn from a GP prior with covariance function $k_{\vartheta}(t,t')$. The aligned sequence, which is unobserved, is given by the same function without the time warp $\mathbf{f}^0_n=f_n(\mathbf{t}^0_{n})$. The joint distribution of the random variables representing the misaligned sequence $\mathbf{f}_{n}$ and the aligned sequence $\mathbf{f}^0_{n}$ according to the prior is
\begin{equation}
 \left[\begin{array}{c} \mathbf{f}_n\\ \mathbf{f}^0_n \end{array}\right] \sim \mathcal{N}\left(\mathbf{0},\left[ \begin{array}{cc} K^{\vartheta}_{\mathbf{t}_n,\mathbf{t}_n} & K^{\vartheta}_{\mathbf{t}_n,\mathbf{t}^0_n}\\ K^{\vartheta}_{\mathbf{t}^0_n,\mathbf{t}_n} & K^{\vartheta}_{\mathbf{t}^0_n,\mathbf{t}^0_n} \end{array}\right] \right)
\end{equation}
The warping function can also be drawn from a GP prior with appropriate smoothness and monotonicity properties. Actually, since only $\mathbf{y}_n$ are observed in the time instants given by $\mathbf{t}_n$, the temporal support $\mathbf{t}^0_n$ of the aligned sequence could be inferred as $\mathbf{t}^0_{n}=g^{-1}_n(\mathbf{t}_{n})$. As the warping function is strictly monotonic and it is bijective, its inverse is well-defined. 

\subsection{Hierarchical Dirichlet Processes and SLDS}
According to Ghahramani and Hinton~\citep{ghahramani_variational_2000}, a {\em switching linear dynamical system} (SLDS) enables modelling of nonlinear dynamics by transitioning between multiple linear dynamical systems (LDS) discretely. Formally:
\begin{equation}
\begin{aligned}
s_{n}  &\sim  \pi ^{(s_{n-1})} \\
    \mathbf{f}_{n} &= A^{(s_{n})} \mathbf{f}_{n-1}  + \omega _n^{(s_{n})} \\
    \mathbf{y}_{n} &= C^{(s_{n})} \mathbf{f}_{n}  + \epsilon _n ^{(s_{n})}.
\label{eq:slds}
\end{aligned}
\end{equation}
Each latent switching state $s_n$ follows a multinomial distribution over $S$ discrete values. These states determine the probabilistic assignment of observations $\mathbf{y}_n$ to their corresponding LDS. It is assumed a first-order Markov dependency between states $s_{n}$ with transitions given by $\pi = \{\pi^{(j)}\}_{j=0}^S$, $\pi^{(j)}\in \mathbb{R}^S_{+}$. Specifically, $\pi^{(0)}$ denotes the initial state distribution and $\pi^{(j)}$ ($j=1,...,S$) denotes the {\em j}th row of the transition matrix. In Equation~\ref{eq:slds} it is assumed that both the dynamics of the latent state process and the observation mechanism depend on the switching state, but a shared observation mechanism can also be assumed, depending on the applications~\citep{fox_2011}.

A major limitation of this framework is assuming that the number of states $S$ is known in advance. Dirichlet processes (DP) are commonly used to build nonparametric Bayesian models, particularly HMMs, such that the number of states is unbounded a priori. The {\em Dirichlet process} $\text{DP}(\gamma, H)$ is a measure on measures, where $\gamma >0$ is a concentration parameter and $H$ is a base probability measure~\citep{ferguson_1973}. We can represent a draw from a Dirichlet process $G_0\sim \text{DP}(\gamma, H)$ as
\begin{equation}
    G_0=\sum_{k=1}^{\infty} \beta_k \delta_{\phi_k},
\end{equation}
where $\phi_k \overset{iid}{\sim} H$ and $\delta _{\phi_k}$ is a probability measure concentrated at $\phi_k$. The weights $\beta_k$ are sampled by means of a {\em stick-breaking process}~\citep{sethuraman_1994}:
\begin{equation}
\beta_k= v_k \prod _{j=1} ^{k-1}(1-v_j), ~~~ v_k\sim \text{Beta}(1,\gamma).
\label{eq:stick_breaking}
\end{equation}
The sequence $\beta = (\beta_k)_{k=1} ^{\infty}$ satisfies $\sum _{k=1} ^{\infty}\beta_k=1$ and it is denoted as $\beta\sim \text{GEM}(\gamma)$. Intuitively, the stick-breaking process involves repeatedly breaking off and discarding a random fraction of a stick that is initially of length 1. This random fraction is, in turn, sampled from a beta distribution. Thus, each weight $v_k$ would define the probability of reaching state $k$, given that the first $k-1$ states were not reached, and each weight $\beta_k$ would define the marginal probability of reaching state $k$.

In a {\em hierarchical Dirichlet process} (HDP), the base measure of a DP is distributed as a DP. Thus, a HDP is defined as a collection of measures $\{G_j\}_{j=1}^\infty$ distributed as a DP with $G_j \overset{iid}{\sim} \text{DP}(\alpha, G_0)$, where $G_0$ is also distributed as a DP with $G_0\sim \text{DP}(\gamma, H)$~\citep{beal_2001,teh_2006}. According to the stick-breaking representation
\begin{equation}
    G_j=\sum_{k=1}^{\infty} \pi_{jk} \delta_{\phi_k},
    \label{eq:gj}
\end{equation}
where each $G_j$ has support at the same points $\phi=(\phi_{k})_{k=1}^{\infty}$ as $G_0$, providing a desirable sharing property for hierarchical clustering. Since each $G_j$ is drawn independently given $G_0$, the weights $(\pi_{jk})_{k=1}^{\infty}$ are independent given $\beta$.

A nonparametric HMM can be based on a set of DPs, one for each value of each state variable. The hierarchical Dirichlet process HMM (HDP-HMM) replaces the set of conditional finite mixture models of the classical HMM with an HDP mixture model. If a finite Bayesian HMM places a Dirichlet($\alpha$) prior on each row of the transition matrix $\pi$, the HDP-HMM considers each of the infinitely many rows of $\pi= \{\pi^{(j)}\}_{j=0}^{\infty}$ to be the stick-breaking pieces of some $G_{j} \sim \text{DP}(\alpha, G_{0})$, this allowing infinite distributions over infinite hidden states. Formally, the parameters of the HDP-HMM are distributed according to
\begin{align}
    \beta & \sim  \text{GEM}(\gamma)\\
    \pi^{(j)} & \overset{iid}{\sim}  \text{DP}(\alpha,\beta) \label{eq:hdp-hmm_pi} \\
    \phi^{(j)} & \overset{iid}{\sim}  H,
\end{align}
where $\beta$ is treated as a density concerning counting measure on $\mathbb{N}$. Each $\pi^{(j)}$ is a draw from a DP and $\mathbb{E}[\pi^{(j)}]=\beta$, thus making the transition distributions concentrate around a similar set of states, providing the desired property of re-using previous states. Both the latent and observed variables in the HDP-HMM are distributed according to
\begin{align}
    s_n  &\sim  \pi ^{(s_{n-1})} \label{eq:hmm_sn} \\
    \mathbf{y}_n | s_n & \sim F(\phi^{(s_n)}), \label{eq:hmm_yn}
\end{align}
where $F(\phi^{(n)})$ is the emission distribution and $\phi^{(n)}$ is the set of emission parameters for the $s_n$th state.

Fox {\em et al.}~\citep{fox_2011} provide an extension of HDP-HMM for modelling SLDS by allowing to switch between an unknown number of linear dynamical systems (HDP-SLDS). Equation (\ref{eq:hmm_yn}) in their approach takes the form of an LDS, and model parameters are distributed as $(A^{(j)},\Sigma _{\omega}^{(j)}, \Sigma _{\epsilon}) \overset{iid}{\sim}  H$, proposing a shared measurement mechanism. Additionally, they provide a {\em sticky transition}, a construction that increases the expected probability of self-transition to prevent unrealistically fast transitions between linear dynamical systems. Authors place priors on the dynamics parameters $A^{(s_{n})}$, $\Sigma _{\omega} ^{(s_{n})}$, and on the measurement noise covariance $\Sigma _{\epsilon}$, and infer their posterior from the data.

\section{The generative model}
\label{sec:generative_model}
\begin{figure}
	\centering
	\includegraphics[width=0.4\linewidth]{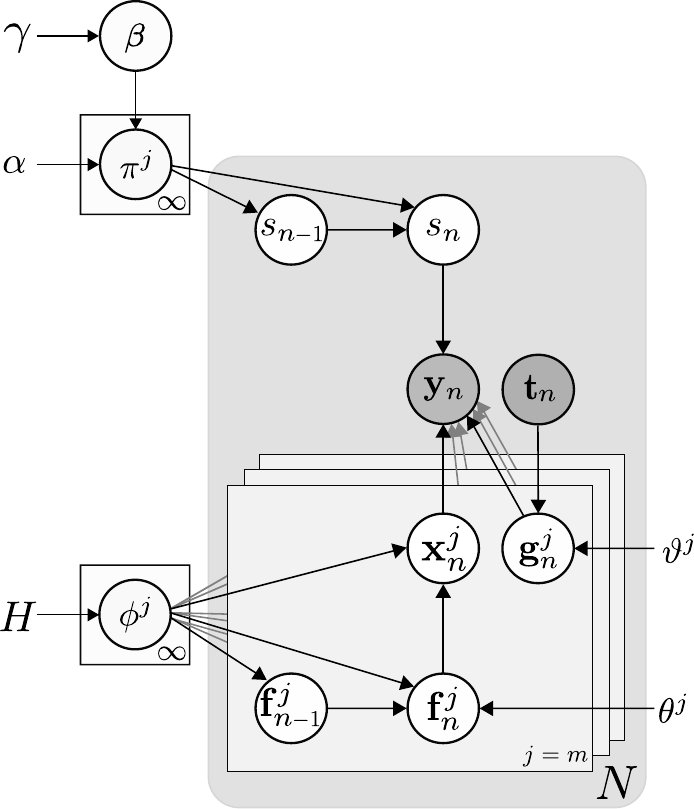}
	\caption{Graphical model. Several linear dynamics are generated to represent the different clusters that evolve across time. Conditioned on a specific setting of the switch variable $s_n=m$, an observation $\mathbf{y}_n$ is generated by LDS $m$ through a linear projection given by equations (\ref{eq:latent_trans}) and (\ref{eq:emission_trans}) from the last observation generated by this same LDS.}
	\label{fig:gen_model}
    \Description[HDP-GPC graphical model]{
    Diagram of the proposed HDP-GPC probabilistic model. A hierarchical Dirichlet process defines global and transition probabilities for the switching states; each state selects cluster-specific Gaussian-process and linear-dynamical components, which generate an observed time-series segment after latent evolution and temporal warping.
    }
\end{figure}

We explain the generative process of the HDP-GP clustering (HDP-GPC) by separating it into a sequence of stages (see Figure~\ref{fig:gen_model}). The first one is a Bayesian nonparametric switching, modelled by an HDP as a prior on a set of HMM transitions, with concentration parameters $\gamma, \alpha >0$:
\begin{align}
    \beta & \sim  \text{GEM}(\gamma),\\
    \pi^{(j)} & \overset{iid}{\sim}  \text{DP}(\alpha,\beta), \label{eq:dp} \\
    \phi^{(j)} & \overset{iid}{\sim}  H_{\text{LDS}},\\
    s_{n}  &\sim  \pi ^{(s_{n-1})}, \label{eq:switching}
\end{align}
where $s_1\sim \pi^{(0)}$, and the parameters corresponding to the linear dynamical system are $(A^{(j)},\Sigma _{\omega}^{(j)}, C^{(j)}, \Sigma _{\epsilon} ^{(j)}) \overset{iid}{\sim}  H_{\text{LDS}}$.

Every time a new state is generated by expression (\ref{eq:switching}) a new LDS is generated by instantiating a GP $f^{(s_n)}(t) \sim \mathcal{GP}(m_{\theta ^{(s_{n})}}(t),k_{\theta ^{(s_{n})}}(t, t'))$, with hyperparameters $\theta ^{(s_{n})}$, on a certain time index set:
\begin{equation}
\mathbf{f} _1\sim \mathcal{N}(\mathbf{0},K^{\theta ^{(s_{n})}}_{\mathbf{t}_1,\mathbf{t}_1}),
\end{equation}
where $\mathbf{t}_1$ denotes the input vector from the first sequence. Since we have no prior knowledge about the mean, we take it to be zero by symmetry. Then, each LDS can evolve a first realisation of the corresponding GP several times as
\begin{align}
    \mathbf{f}^{(s_{n})}_{n} &= A^{(s_{n})} \mathbf{f}^{(s_{n})}_{n-1} + \omega _n^{(s_{n})}, \label{eq:latent_trans}\\
    \mathbf{x}^{(s_{n})}_{n} &= C^{(s_{n})} \mathbf{f}^{(s_{n})}_{n} + \epsilon ^{(s_{n})} _n. \label{eq:emission_trans}
\end{align}
Thus, each LDS represents a certain morphology as a GP which evolves across time, providing new instances that are continually being added to the chain of observations. Note that since (\ref{eq:latent_trans}) and (\ref{eq:emission_trans}) are composed of linear transformations, the GP structure is preserved. Actually, a pseudo-observation $\mathbf{x}^{(s_{n})}_n$ is generated at each step, which is then affected by an irregular time delay before it is finally observed. We model this by assuming $\mathbf{x}^{(s_{n})}_n$ as a finite representation at the input points of the GP resulting from the projection of $\mathbf{f}^{(s_{n})}_n$ by equations (\ref{eq:latent_trans}) and (\ref{eq:emission_trans}). We recall that equation (\ref{eq:gaussian_process}) allows us to predict $x^{(s_n)}(t)\sim \mathcal{GP}(m_{x^{(s_n)}}(t),k_{x^{(s_n)}}(t,t'))$ on new different test points, particularly, on those given by the time delay. The time delay produces a temporal misalignment that can be modeled by a monotonic warping given by a GP on the time index set $g^{(s_n)}(t) \sim  \mathcal{GP}(m_{\vartheta ^{(s_{n})}}(t),k_{\vartheta ^{(s_{n})}}(t,t'))$, with hyperparameters $\vartheta ^{(s_{n})}$, leading to a new temporal support of the observations that, in the absence of noise, is given by 
\begin{equation}
 \mathbf{t}_n^w \triangleq \mathbf{g}^{(s_{n})}_n\sim \mathcal{N}(m_{\vartheta ^{(s_{n})}}(\mathbf{t}_n), K^{\vartheta ^{(s_{n})}}_{\mathbf{t}_n,\mathbf{t}_n}).
\end{equation}
Both the target observations $\mathbf{y}_n$, inferred at the new locations $\mathbf{t}^w_n$, and $\mathbf{x}^{(s_{n})}_n$ have joint distribution under the prior
\begin{equation}
 \left[\!\!\begin{array}{c} \mathbf{x}^{(s_{n})}_n\\ \mathbf{y}_n \end{array}\!\!\right] \sim \mathcal{N}\left( \left[\!\begin{array}{c} m^{\varphi ^{(s_n)}}_{\mathbf{t}_n}\\  m^{\varphi ^{(s_n)}}_{\mathbf{t}^w_n} \end{array}\!\right],\left[ \begin{array}{cc} K^{\varphi ^{(s_n)}}_{\mathbf{t}_n,\mathbf{t}_n} & K^{\varphi ^{(s_n)}}_{\mathbf{t}_n,\mathbf{t}^w_n}\\ K^{\varphi ^{(s_n)}}_{\mathbf{t}^w_n,\mathbf{t}_n} & K^{\varphi ^{(s_n)}}_{\mathbf{t}^w_n,\mathbf{t}^w_n} \end{array}\right] \right) \label{eq:joint_dist_x}
\end{equation}
where 
\begin{align}
m^{\varphi ^{(s_n)}}_{\mathbf{t}} & =K^{\theta ^{(s_{n})}}_{\mathbf{t},\mathbf{t}_n}K^{\theta ^{(s_{n})}-1}_{\mathbf{t}_n,\mathbf{t}_n}\mathbb{E}[\mathbf{x}^{(s_{n})}_n], \nonumber \\
K^{\varphi ^{(s_n)}}_{\mathbf{t},\mathbf{t}'} &=K^{\theta ^{(s_{n})}}_{\mathbf{t},\mathbf{t}'}-K^{\theta ^{(s_{n})}}_{\mathbf{t},\mathbf{t}_n}K^{\theta ^{(s_{n})}-1}_{\mathbf{t}_n,\mathbf{t}_n}K^{\theta ^{(s_{n})}}_{\mathbf{t}_n,\mathbf{t}'} + K^{\theta ^{(s_{n})}}_{\mathbf{t},\mathbf{t}_n}K^{\theta ^{(s_{n})}-1}_{\mathbf{t}_n,\mathbf{t}_n}\cov(\mathbf{x}_n^{(s_n)})K^{\theta ^{(s_{n})}-1}_{\mathbf{t}_n,\mathbf{t}_n}K^{\theta ^{(s_{n})}}_{\mathbf{t}_n,\mathbf{t}'}.
\end{align}
Ultimately, a vector of observations $\mathbf{y}_n$ is generated by marginalization
\begin{equation}
\mathbf{y}_n \sim \mathcal{N}(m^{\varphi ^{(s_n)}}_{\mathbf{t}^w_n}, K^{\varphi ^{(s_n)}}_{\mathbf{t}^w_n,\mathbf{t}^w_n})
\end{equation}

\section{Variational inference}
\label{sec:variational_inference}
We now focus on estimating the posterior probabilities of the hidden variables of the model $\mathcal{M}$ conditional on the set of observed variables given as a set of time sequences $\mathcal{Y}=\{\mathbf{y}_1,...,\mathbf{y}_N\}$. Our goal is therefore to infer the posterior distribution $p(\mathcal{M}|\mathcal{Y})$ where $\mathcal{M}$ denotes the collection of variables $\mathcal{M} = \{\beta, \pi^{j}, \phi^{j}, s_{n}, \mathbf{f}_{n},  \mathbf{x}_{n}, \mathbf{g}_{n}\}$ with $j=1,\dots,\infty$ and $n=1,\dots, N$. From Bayes' theorem
\begin{equation}
p(\mathcal{M}|\mathcal{Y}) \propto p(\mathcal{Y}|\mathcal{M}) p(\mathcal{M}),
\end{equation}
When conditioning on the observations, two main sources of intractability can be identified in computing the posterior: first, the switching variable induces a coupling between all linear dynamics, leading to an exponential growth of the Gaussian terms involved in the posterior distribution~\citep{ghahramani_variational_2000}; second, as in many hierarchical Bayesian models, the posterior in HDP is not available in closed form~\citep{teh_2006}. Previous approaches to infer the posterior for the HDP-HMM model have used Gibbs sampling~\citep{fox_2011}, showing slow convergence and limited scalability, or stochastic variational inference~\citep{johnson_stochastic_2014}, easily scaling to massive data but leading to poor accuracy as the latent space dimension increases~\citep{dhaka_2020}. Following~\citep{hughes_2015}, we adopt a scalable variational inference approach that allows us to exploit the sequential structure of the model and hence is easily applicable to streaming data. 

Variational inference turns the inference problem into an optimisation problem~\citep{bishop_pattern_2006, blei_2017}, whereby we approximate the intractable true posterior with the distribution from a tractable family $\mathcal{Q}$ that minimises the Kullback-Leibler (KL) divergence to the true posterior $q^{*}=\mathrm{argmin}_{q\in\mathcal{Q}} \mathrm{KL}(q||p)$. Since $\log p(\mathcal{Y})=\mathcal{L}(q)+\mathrm{KL}(q||p)$, we can equivalently seek $q^*$ to maximise the lower bound of the log marginal evidence: 
\begin{equation}
\mathcal{L}(q) = \mathbb{E}_{q}\left[\log p(\mathcal{M},\mathcal{Y}) - \log q(\mathcal{M})\right]
\label{eq:elbo}
\end{equation}
As stated in~\citep{saul_1995}, a natural approach is to consider a {\em mean-field variational family} that factors according to tractable substructures where exact inference algorithms can be used. We propose the factorized distribution illustrated in table~\ref{tab:variational_factors} with $q(\mathcal{\cdot})=q(\beta)q(\pi)q(\phi)q(\mathcal{S})q(\mathcal{F}, \mathcal{X})q(\mathcal{G})$, with global factors $q(\beta)$, $q(\pi)$, $q(\phi)$, and local factors $q(\mathcal{S})$, $q(\mathcal{F}, \mathcal{X})$, $q(\mathcal{G})$, where $\mathcal{S}=\{s_1,...,s_N\}$, $\mathcal{F}=\{\mathbf{f}_1,...,\mathbf{f}_N\}$, $\mathcal{X}=\{\mathbf{x}_1,...,\mathbf{x}_N\}$ and $\mathcal{G}=\{\mathbf{g}_1,...,\mathbf{g}_N\}$. While the objective of optimising (\ref{eq:elbo}) is not convex, a local maximum can be achieved by a coordinate ascent procedure, where we iteratively optimise each factor while holding the others fixed~\citep{bishop_pattern_2006}. For each of the factors $\mathbf{z}_i\in \mathcal{M}$, the optimal $q_i(\mathbf{z}_i)$ is then proportional to the exponentiated expected log of the joint distribution
\begin{equation}
q^{*}_j(\mathbf{z}_j) \propto \operatorname{exp}\{\mathbb{E}_{i\neq j} [\log p(\mathcal{M},\mathcal{Y})]\}
\end{equation}

\begin{table}
\centering{
\footnotesize
\setlength{\tabcolsep}{4pt}
\renewcommand{\arraystretch}{1.15}
\caption{Variational factors used to approximate the posterior distribution of HDP-GPC.}
\label{tab:variational_factors}
\begin{tabular}{l l l l}
\hline\hline
Factor & Latent variables & Variational family & Role in inference \\ \hline

$q(\beta)$
&
Stick-breaking variables
&
$\displaystyle \prod_{k=1}^{K} 
\operatorname{Beta}\!\left(
v_k | \lambda_k\eta_k,(1-\lambda_k)\eta_k
\right)$
&
\begin{tabular}[c]{@{}l@{}}
Controls the global HDP weights $\boldsymbol{\beta}$ \\
and penalises unused clusters.
\end{tabular}
\\

$q(\pi)$
&
Transition distributions
&
$\displaystyle \prod_{j=0}^{K}
\operatorname{Dir}
\left(
\pi^{(j)}
|
\kappa_{j1},\ldots,\kappa_{j,K+1}
\right)$
&
\begin{tabular}[c]{@{}l@{}}
Updates Markov transition probabilities \\
from expected transition counts.
\end{tabular}
\\

$q(\mathcal{S})$
&
Switching states
&
$\displaystyle
\prod_{n=1}^{N}
\prod_{j=1}^{K}
\prod_{k=1}^{K}
\xi_{njk}^{s_{n-1,j}s_{nk}}
$
&
\begin{tabular}[c]{@{}l@{}}
Provides cluster responsibilities \\
and expected transition counts.
\end{tabular}
\\

$q(\mathcal{G})$
&
Warping functions
&
$\displaystyle
\prod_{n=1}^{N}
\prod_{k=1}^{K}
\delta_{\widehat{\mathbf{g}}^{(k)}_n}
\!\left(
\mathbf{g}^{(k)}_n
\right)
$
&
\begin{tabular}[c]{@{}l@{}}
Approximates temporal alignment \\
by a MAP monotone warp.
\end{tabular}
\\

$q(\phi)$
&
LDS parameters
&
$\displaystyle
\prod_{k=1}^{K}
q\left(\mathbf{A}^{(k)},\boldsymbol{\Sigma}^{(k)}_\omega\right)
q\left(\mathbf{C}^{(k)},\boldsymbol{\Sigma}^{(k)}_\epsilon\right)
$
&
\begin{tabular}[c]{@{}l@{}}
Learns cluster-specific evolution \\
and emission parameters.
\end{tabular}
\\

$q(\mathbf{A}^{(k)},\boldsymbol{\Sigma}^{(k)}_\omega)$
&
Latent transition parameters
&
$\mathcal{MN IW}
\left(
\widetilde{\mathbf{M}}^{(k)}_A,
\widetilde{\mathbf{W}}^{(k)}_A,
\widetilde{\mathbf{V}}^{(k)}_A,
\widetilde{\nu}^{(k)}_A,
\widetilde{\mathbf{S}}^{(k)}_\omega
\right)$
&
\begin{tabular}[c]{@{}l@{}}
Conjugate update for the \\
LDS transition model.
\end{tabular}
\\

$q(\mathbf{C}^{(k)},\boldsymbol{\Sigma}^{(k)}_\epsilon)$
&
Emission parameters
&
$\mathcal{MN IW}
\left(
\widetilde{\mathbf{M}}^{(k)}_C,
\widetilde{\mathbf{W}}^{(k)}_C,
\widetilde{\mathbf{V}}^{(k)}_C,
\widetilde{\nu}^{(k)}_C,
\widetilde{\mathbf{S}}^{(k)}_\epsilon
\right)$
&
\begin{tabular}[c]{@{}l@{}}
Conjugate update for the \\
emission transformation.
\end{tabular}
\\

$q\left(\mathcal{F},\mathcal{X}\right)$
&
Latent and emission processes
&
$\displaystyle
\prod_{k=1}^{K}
q\left(\mathcal{X}^{(k)}|\mathcal{F}^{(k)}\right)
q
\left(\mathcal{F}^{(k)}\right)
$
&
\begin{tabular}[c]{@{}l@{}}
Computes posterior GP/LDS representatives \\
by responsibility-weighted Kalman smoothing.
\end{tabular}
\\ \hline\hline
\end{tabular}}
\end{table}

A common approach to approximate the posterior in HDP models is to use a truncated variational distribution~\citep{bryant_2012}, such that, for a truncation parameter $K$ representing the active clusters, we fix $q(s_n)=0$ for any $s_n>K$. Thus, we can initialise with a small $K$ and let the inference discover new clusters as it progresses. Truncation only affects the form of local factors; global factors with indices greater than $K$ are conditionally independent of the observations and only differ from their prior under $p(\cdot)$ for the first $K$ elements.

{\em Factor} $q(\mathcal{S})$. We propose the usual factorisation for discrete Markov chains~\citep{ghahramani_variational_2000,beal_2003}. We are interested in the functional dependence of the generative model on the variable $\mathcal{S}$, so the log of the optimised factor is given by
\begin{align}
\log q^*(\mathcal{S}) &=\mathbb{E}_{\pi}[\log p(\mathcal{S}|\pi)] + \mathbb{E}_{\mathcal{X},\mathcal{G}} [\log p(\mathcal{Y}|\mathcal{S},\mathcal{X},\mathcal{G})]+\mathrm{const.}  \nonumber \\
&= \sum^{N}_{n=1}\sum _{j=1}^{K} \sum _{k=1}^{K} s_{n-1,j} s_{nk} \log \zeta_{njk}+ \mathrm{const.}
\label{eq:q_S}
\end{align}
where $s_{nk}=1$ if the switching variable is in state $k$ at time $n$ and $0$ otherwise. The initial term for $n=0$ is defined as $s_{0k}\zeta_{00k}$ with $\zeta_{00k} = \mathbb{E}[\log \pi_{0k}]$ acting as an initial state distribution. For $n>0$, 
\begin{align}
\log \zeta_{njk} &= \mathbb{E}[\log \pi_{jk}] + \mathbb{E}[\log p(\mathbf{y}_{n}|\mathbf{x}_n^{(s_{n})},\mathbf{g}^{(s_{n})}_n, s_{n})], \label{eq:zeta}
\end{align}
the full derivation can be found in Appendix~\ref{ap:variational_updates}. We then use $\zeta_{njk}$ to perform forward-backward recursion to compute $\xi_{njk} \triangleq q(s_{n-1}=j, s_{n}=k)$. The optimised factor adopts the same form as the prior distribution $p(\mathcal{S}|\pi)$, as expected. The quantities 
\begin{equation}
\mathbb{E}[s_{nk}]=\sum^K_{j=1} \xi_{njk}=r_{nk},
\label{eq:responsibilities}
\end{equation}
play the usual role of responsibilities and can be interpreted as the expected number of times that state $k$ is visited at time $n$, as it is common in HMM literature. Moreover, the quantities $\xi_{njk}$ can be interpreted as the probability of being in state $j$ at time $n-1$, and state $k$ at time $n$, as it is common in HMM literature. Following~\citep{bishop_pattern_2006}, we define some statistics weighted by the responsibilities: $N_k=\sum_{n=1}^N r_{nk}$, and $N_{jk}=\sum _{n=1}^N \xi_{njk}$.

{\em Factor} $q(\beta)$. The update to $q(\beta)$ is not conjugate given the other factors. The usual approach for $q(\beta)$, a simple point estimate $q(\beta)=\delta_{\beta^*}(\beta)$~\citep{johnson_stochastic_2014, fox_hdp-hmm_2008}, has proven not to penalise empty clusters~\citep{hughes_2015}, and it is suggested instead a proper Beta distribution over each stick breaking weight $v_k$:
\begin{equation}
q^*(v|\lambda,\eta)=\prod ^{\infty}_{k=1} \operatorname{Beta}(v_k|\lambda _k\eta_k,(1-\lambda_k)\eta_k), \label{eq:param_stick_breaking}
\end{equation}
where the hyperparameter $\lambda_k \in (0,1)$ establishes the mean of $v_k$ and $\eta_k>0$ controls its variance.

{\em Factor} $q(\pi)$. It is convenient to rewrite $\pi^{(j)}$ as a finite partition $[\pi_{j1},...,\pi_{jK},\pi_{jK+1}]\sim \operatorname{Dir}(\pi_j|\alpha\beta_1,...,\alpha\beta_{K+1})$, where $\pi_{jK+1}\triangleq \sum^{\infty}_{h=K+1}\pi_{jh}$ represents the aggregate mass of all inactive clusters and $\beta_{K+1}\triangleq \sum^{\infty}_{h=K+1}\beta_{h}$. The log of the optimised factor is given by 
\begin{align}
\log q^*(\pi) &= \mathbb{E}_{\mathcal{S}}[\log p(\mathcal{S}|\pi)] + \mathbb{E}_{v}[\log p(\pi|v)] + \mathrm{const.} \nonumber \\
&= \sum_{j=1}^{K+1} \sum_{k=1}^{K+1}\left[\sum_{n=1}^N \xi_{njk}+\alpha \mathbb{E}_v[\beta_k] -1 \right] \log \pi_{jk} + \mathrm{const.}
\label{eq:q_pi}
\end{align}
We recognise a product of Dirichlet distributions with updated parameters 
\begin{equation}
q^*(\pi) = \prod_{j=1}^{K+1} \operatorname{Dir} (\pi_j|\kappa_{j1},...,\kappa_{jK+1}), 
\label{eq:kappa}
\end{equation}
where $\kappa_{jk}=\alpha \mathbb{E}_v[\beta_k]+N_{jk}$ and a term $N_{jk}$ is added to the prior, providing the expected number of times that the observations are generated after a transition $\pi_{jk}$. Note that $N_{jK+1}=0$ since the inactive clusters have no elements. 

{\em Factor} $q(\mathcal{G})$. The log of the optimised factor is given by
\begin{equation}
\log q^*(\mathcal{G}) = \log p(\mathcal{G}) + \mathbb{E}_{\mathcal{S},\mathcal{X}}[\log p(\mathcal{Y}|\mathcal{S},\mathcal{X},\mathcal{G})] + \mathrm{const.}
\label{eq:g}
\end{equation}
The second term at the right-hand side of (\ref{eq:g}) involves a composition of GPs $(x \circ g)$, which does not have a closed-form solution. Following~\citep{lawrence_2007}, we use a {\em maximum a posteriori} approach, where we have assumed an independent time warping affecting each time sequence for each different cluster.  Monotonicity is mandatory for a warping function, and we adopt a parametrisation of each $\mathbf{g}^k_n$ by using a set of auxiliary variables and appropriate hyperparameters to ensure smoothness.

{\em Factor} $q(\phi)$. As stated in~\citep{west_1997}, the matrix-normal inverse-Wishart prior provides conjugates to the likelihood equations  (\ref{eq:latent_trans}) and (\ref{eq:emission_trans}), which can be interpreted as linear regression problems that can be independently solved for each cluster. For convenience, we deal with the two components of the LDS separately, assuming the usual prior
\begin{align}
p(A^k,\Sigma^k_\omega) &=p(A^k|\Sigma^k_\omega)p(\Sigma^k_\omega) \nonumber \\
&= \mathcal{MN}(M^k_A,W^k_A,V^k_A)\mathcal{IW}(\nu_{0},S^k_\omega).
\end{align}
The subsequent updates are derived in Appendix~\ref{ap:variational_updates}.

{\em Factor} $q(\mathcal{X},\mathcal{F})$. 
The variational posterior distribution $q^*(\mathcal{X},\mathcal{F})$ can be written in the form $q^*(\mathcal{X},\mathcal{F})=q^*(\mathcal{X}|\mathcal{F})q^*(\mathcal{F})$. The log of the optimised factor is given by
\begin{align}
\log q^*(\mathcal{X},\mathcal{F}) &= \mathbb{E}_{\phi}[\log p(\mathcal{X}|\mathcal{F},\phi)] + \mathbb{E}_{\phi}[\log p(\mathcal{F}|\phi)] + \mathbb{E}_{\mathcal{S},\mathcal{G}} [\log p(\mathcal{Y}|\mathcal{S},\mathcal{X},\mathcal{G})] + \mathrm{const.}
\end{align}
A standard message passing algorithm can be applied to each cluster to compute the filtering posterior messages $p(\mathbf{f}_{n}|\{\mathbf{y}_{i}\}^{n}_{i=1})$ and the smoothing messages $p(\mathbf{f}_{n}|\{\mathbf{y}_{i}\}^{N}_{i=1})$. Which later define the predictive distribution used to estimate the emission distribution for the rest of the factors. The full derivation of the forwards-backwards equations is provided in Appendix~\ref{ap:variational_updates}. 

Let us notice that the posterior distribution $q^{*}(\mathbf{x}_n^k$) can show heteroscedastic behaviour, in so far as previous observations in the same LDS lead to a non-diagonal covariance matrix, as will be seen later on with real examples.

\subsection{Variational objective}
The variational objective can be decomposed as
\begin{equation}
\mathcal{L} =\mathcal{L}_{OBS} + \mathcal{L}_{HDP} + \mathcal{L}_{W}+ \mathbb{H}[q(\mathcal{S}\,)].
\label{eq:lower_bound}
\end{equation}
We use equation (\ref{eq:lower_bound_ap}) to check convergence in the coordinate ascent procedure. A full derivation of the variational objective can be found in the supplementary material. 

\subsection{Variational algorithm}
The optimisation of the evidence lower bound involves an iterative re-estimation procedure following a bottom-up strategy from local to global variables. It is important to note that the initial model $p(\mathcal{M})$ can be a completely uninformed model or a pre-trained model.

\subsection{Offline variational algorithm}
The algorithm iterates through the entire dataset $\mathcal{Y}$, ensuring the convergence of the local step before updating the global variational factors. The truncation level $K$ varies during inference until variational lower bound convergence, as stated in the Algorithm~\ref{alg:offline_algorithm}.

\begin{algorithm}
\renewcommand{\algorithmicrequire}{\textbf{Input:}}
\renewcommand{\algorithmicensure}{\textbf{Output:}}
\caption{Offline variational inference}\label{alg:offline_algorithm}
\begin{algorithmic}
\REQUIRE A dataset $\mathcal{Y}=\{(\mathbf{t}_n,\mathbf{y}_n)\}_{n=1}^N$ of time sequences
\REQUIRE A model $p(\mathcal{M},\mathcal{Y})$
\ENSURE The variational density $q(\mathcal{M}|\mathcal{Y})$
\WHILE{$\mathcal{L}$ has not converged}
\WHILE{$\mathcal{L}_{OBS}$ has not converged}
\FORALL {$k=1,2,...$}
\STATE // local update
\FORALL {$n=1,...,N$}
\STATE compute $q(s_{nk})$ using eq. \eqref{eq:zeta}
\ENDFOR
\FORALL{$m=1,...,n_k$}
\STATE compute $q(\mathbf{g}_m^k)$ numerically using eq. \eqref{eq:maximizing_g}
\STATE compute $q(\mathbf{f}_{m}^{k})$ and $q(\mathbf{x}_{m}^{k})$ using eq. \eqref{eq:forward_backward}
\ENDFOR
\STATE compute $q(\phi^{k})$ using eq. \eqref{eq:dynamical_param_estimators}
\ENDFOR
\ENDWHILE
\FORALL {$k=1,2,...$}
\STATE // global update
\STATE compute $q(\pi_{k})$ using eq. \eqref{eq:kappa}
\STATE compute $q(v_{k})$ numerically using eq. \eqref{eq:L_HDP}
\ENDFOR
\STATE compute $\mathcal{L} = \mathcal{L}_{OBS} + \mathcal{L}_{HDP} + \mathcal{L}_{W} + \mathbb{H}[q(\mathcal{S})]$
\ENDWHILE
\RETURN $q(\mathcal{M}|\mathcal{Y})$
\end{algorithmic}
\end{algorithm}

\subsection{Online variational algorithm}
The algorithm handles each instance of the dataset $\mathcal{Y}$ in an ordered fashion. Using only one sample at a time limits the computation of the variational objective, resulting in unbiased noisy estimates, which do not guarantee an improvement over the variational bound evaluated using the entire dataset. In return, a single pass over all the data is needed, as stated in Algorithm~\ref{alg:online_algorithm}.

\begin{algorithm}
\renewcommand{\algorithmicrequire}{\textbf{Input:}}
\renewcommand{\algorithmicensure}{\textbf{Output:}}
\caption{Online variational inference}\label{alg:online_algorithm}
\begin{algorithmic}
\REQUIRE A dataset $\mathcal{Y}=\{(\mathbf{t}_n,\mathbf{y}_n)\}_{n=1}^N$ of time sequences
\REQUIRE A model $p(\mathcal{M},\mathcal{Y})$
\ENSURE The variational density $q(\mathcal{M}|\mathcal{Y})$
\FORALL {$(\mathbf{t}_n,\mathbf{y}_n)\in \mathcal{Y}$}
\WHILE{$\mathcal{L}$ has not converged}
\FORALL {$k=1,2,...$}
\STATE // local update
\STATE compute $q(\mathbf{g}_n^k)$ numerically using eq. \eqref{eq:maximizing_g}
\STATE compute $q(s_{nk})$ using eq. \eqref{eq:zeta}
\FORALL {$m=1,...,n_k$}
\STATE compute $q(\mathbf{f}_{m}^{k})$ and $q(\mathbf{x}_{m}^{k})$ using eq. \eqref{eq:forward_backward}
\ENDFOR
\STATE // global update
\STATE compute $q(\pi_{k})$ using eq. \eqref{eq:kappa}
\STATE compute $q(v_{k})$ numerically using eq. \eqref{eq:L_HDP}
\STATE compute $q(\phi^{k})$ using eq. \eqref{eq:dynamical_param_estimators}
\ENDFOR
\STATE compute $\mathcal{L} = \mathcal{L}_{OBS} + \mathcal{L}_{HDP} + \mathcal{L}_{W} + \mathbb{H}[q(\mathcal{S})]$
\ENDWHILE
\ENDFOR
\RETURN $q(\mathcal{M}|\mathcal{Y})$
\end{algorithmic}
\end{algorithm}

\subsection{Multi-output extension}
\label{subsec:multi-output}
In many applications, observations are vector-valued, commonly resulting from multiple information sources, e.g. multi-lead ECG~\citep{Wagner_2022}, multi-lead electroencephalogram (EEG)~\citep{Guttag_2010}, human motion recognition~\citep{Anguita_2013}, industrial process monitoring~\citep{Zhou_2023}, or directional wave analysis~\citep{Hasselmann_1980}. From a regression point of view, a multi-output problem is characterised by an input domain (usually time or space) and a vector-valued observation $\mathbf{y}_{n}^{(l)}$ with $l=\{1,\dots, L\}$ where $l$ indicates the distinct output channel or task.

The HDP-GPC generative model could be adapted to these requirements by letting each cluster emit a multi-output Gaussian process (MOGP)~\citep{Alvarez_2012}. However, once the LDS update is introduced (Equations~\eqref{eq:latent_trans}-\eqref{eq:emission_trans}), output channels become entangled through the latent function: a fully coupled formulation requires block-structured LDS parameters ($A, C, \Sigma_{\omega}, \Sigma_{\epsilon}$) that explicitly model cross-output correlations, and the corresponding filtering/smoothing updates must jointly account for these correlations.

We adopted a computationally simple alternative: we assume conditional independence across outputs given the discrete regime, and we enforce a shared assignment $s_{n}$ across outputs to ensure clustering remains stable across channels. This corresponds to modelling each output with its own SLDS-GP emission, but coupling them through the single shared discrete state sequence. It is important to note that the model runs only one HDP-HMM, as $s_{n}$ is shared, and the emission evidence is aggregated in a weighted fashion across outputs. Let $\{\mathbf{y}_{n}^{(l)}\}_{l=1}^{L}$ denote the $L$ output channels for observation $n$, and let $w_{n}^{(l)} \geq 0$ be weights satisfying $\sum_{l=1}^{L} w_{n}^{(l)}=1$. Let us assume $s_{n}$ fixed, then instead of introducing a multi-output covariance, we weight the emission evidence per output through a composite likelihood:
\begin{equation}
    p(\mathbf{y}_{n}\, |\, s_{n}, \mathcal{X}, \mathcal{G}) \propto \prod_{l=1}^{L}\left[ p(\mathbf{y}_{n}^{(l)}\,|\,s_{n}, \mathcal{X}\,^{(l)}, \mathcal{G}^{(l)})\right]^{w_{n}^{(l)}},
\end{equation}
preserving a shared discrete state across outputs while avoiding the cost of a fully coupled multi-output GP/LDS covariance. The choice of $w_{n}^{(l)}$ is left to the application and will be specified in the following chapters.

Considering the variational updates, the definition of the weighted inputs only affects directly in the second term of equation~\eqref{eq:zeta}, which now takes the form of a weighted sum
\begin{equation}
    \sum_{l=1}^{L} w_{n}^{(l)} \mathbb{E}[\log p(\mathbf{y}_{n}^{(l)}|\mathbf{x}_n^{(l,s_{n})},\mathbf{g}^{(l,s_{n})}_n, s_{n})]
\end{equation}
where is computed exactly as in equation~\eqref{eq:zeta} but using the corresponding output-specific emission covariance for each output channel $l$ of $\mathbf{y}_{n}^{(l)}$. In addition, an outer loop over outputs has to be added in algorithms~\ref{alg:offline_algorithm} and~\ref{alg:online_algorithm} which isolates the assignation computation after LDS pass for each output.

\section{Hyperparameter initialization}
\label{sec:hyperparameters}
Hyperparameters refer to variables that are not estimated in the variational inference process. These variables play a key role in terms of both the behaviour and performance of the model. The main aim of this section is to explore the impact of these hyperparameters on cluster formation and evolution, ultimately affecting the number of inferred clusters. This analysis is intended to aid in fine-tuning the model to achieve the desired performance.

The HDP is controlled by two concentration parameters. The parameter $\gamma$ controls the number of effective clusters. A larger $\gamma$ implies a slower decay of stick-breaking weights $\beta_1, \beta_2, \beta_3,...$ and thus more clusters. The parameter $\alpha$ controls the similarity of the transition distribution from each state to the $\beta$ distribution.

Each LDS is controlled by the parameters of the corresponding matrix-normal inverse-Wishart priors, both for the evolution of $\mathbf{f}^k_n$ and $\mathbf{x}^k_n$. Initialisation of these priors takes place whenever a new cluster is produced, hence we set $M_A=I$ and $M_C=I$, so that each new LDS will be provided with loose constraints at its onset.

\begin{figure}[h!]
\centering
    \includegraphics[width=0.35\columnwidth]{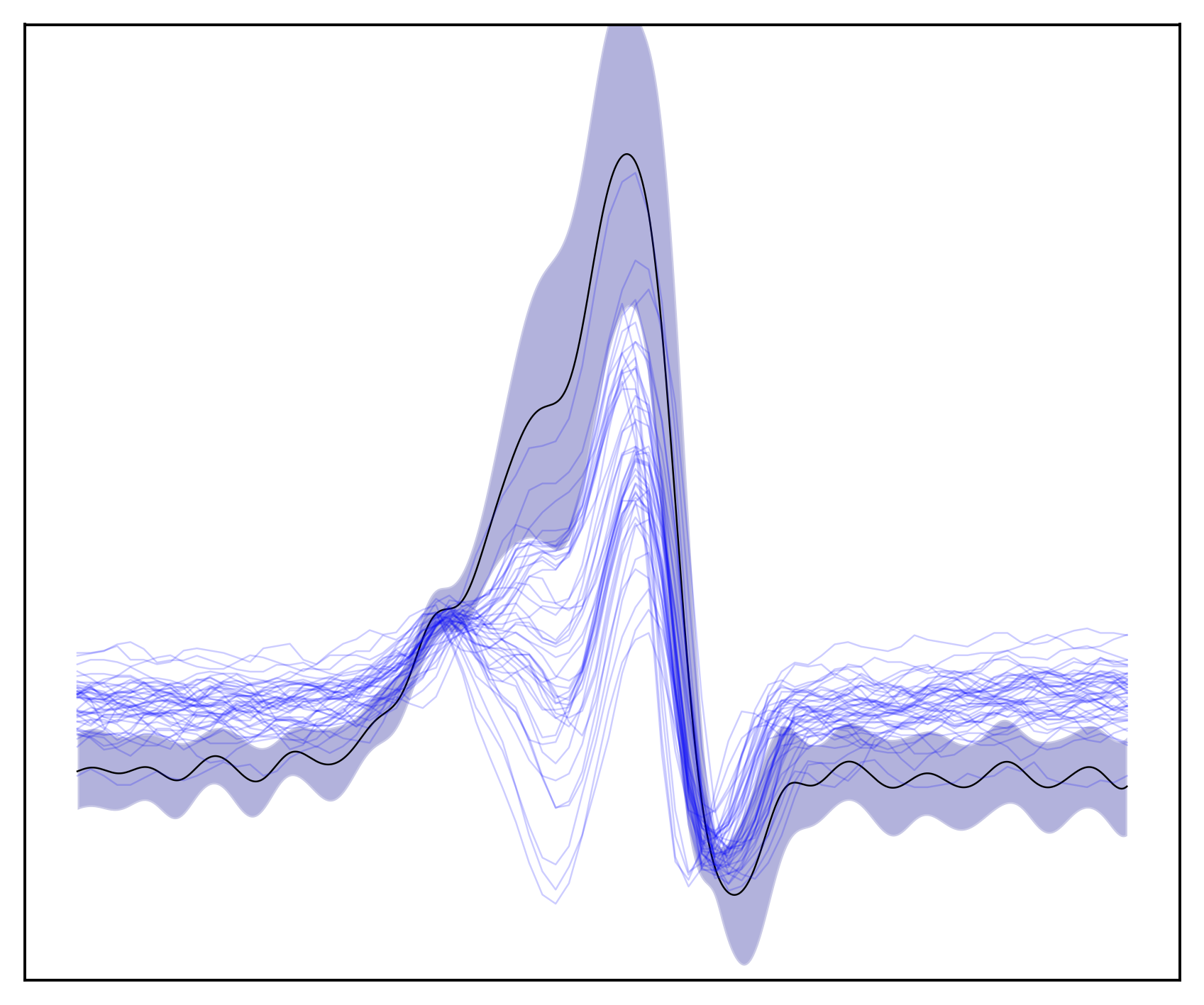}
    \includegraphics[width=0.35\columnwidth]{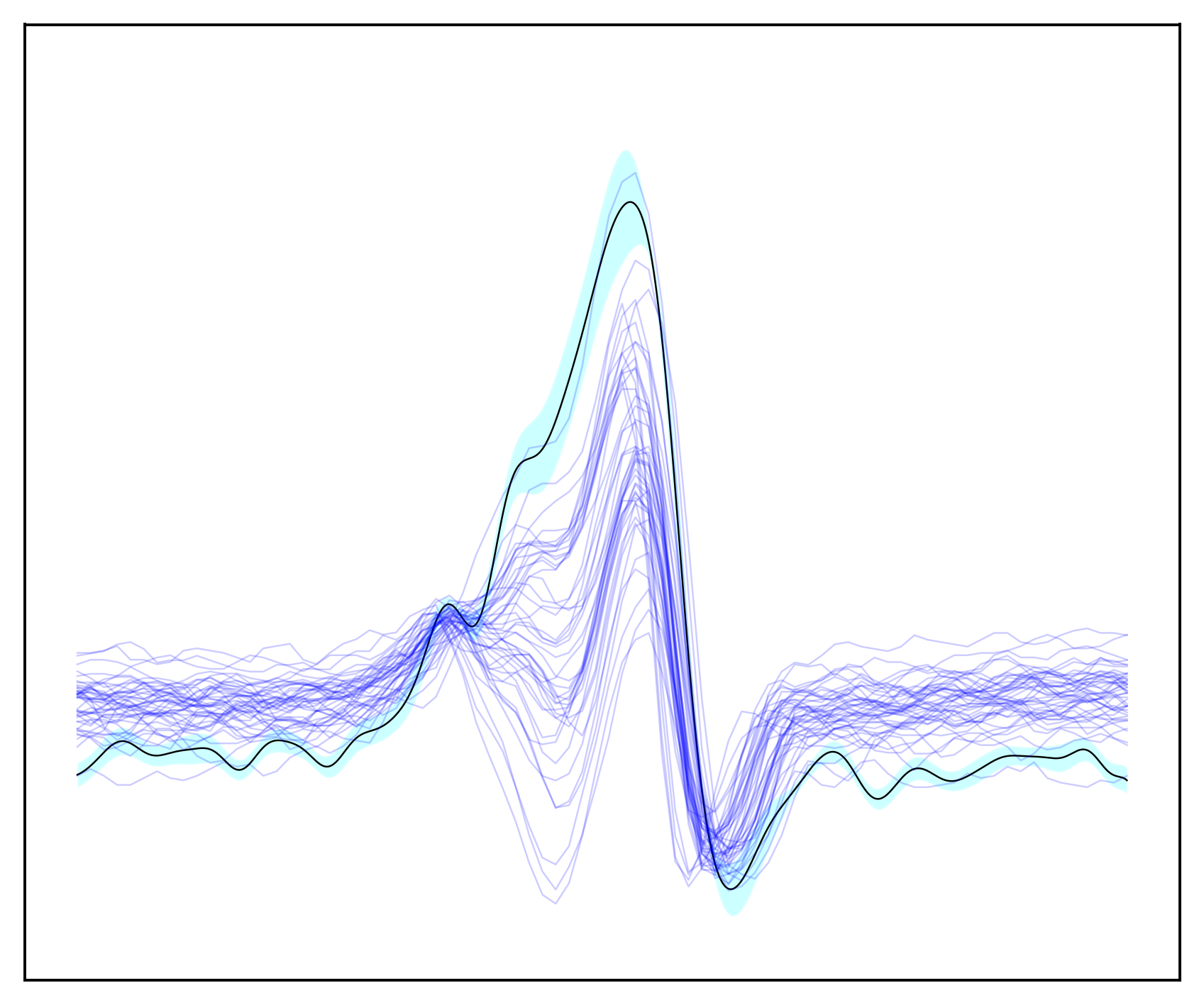}
    \caption{Model enhancing plasticity. A total of 50 heartbeats are represented, manually annotated as Normal, from record 114 of the MIT-BIH Arrhythmia Database [11:30.000,12:20.000]. The resulting GP $f(t)$ (left) and $x(t)$ (right), after evolving along with these sequences, were sampled over a dense time axis. The mean of the posterior is shown as a black line. A 95\% confidence region for the posterior is shown in cyan blue. Kernel hyperparameters for $k_{\theta}(t,t')$ are $\sigma^{2}_{f} = 16.0^{2}$, $\ell = 2.5$, $\sigma^{2}_{n} = 5.0^{2}$. LDS priors: $S_{\omega} = 10.0^{2}I, S_{\epsilon} = 5.0^{2}I$, where $I$ is the identity matrix. For clarity purposes, no warping function has been used. This model embraces the totality of the heartbeats in a single cluster.}
    \label{fig:plasticity_model}
    \Description[Plastic GP model for ECG beats]{
    Two-panel ECG plot showing fifty normal heartbeat segments together with posterior Gaussian-process representatives under a plastic configuration. The posterior mean adapts to the evolving morphology, and the uncertainty band of the emission step remains small relative to the variance of the observations indicating a plastic behaviour of the model adapting to new evidence.
    }
\end{figure}

\begin{figure}[h!]
\centering 
    \includegraphics[width=0.35\columnwidth]{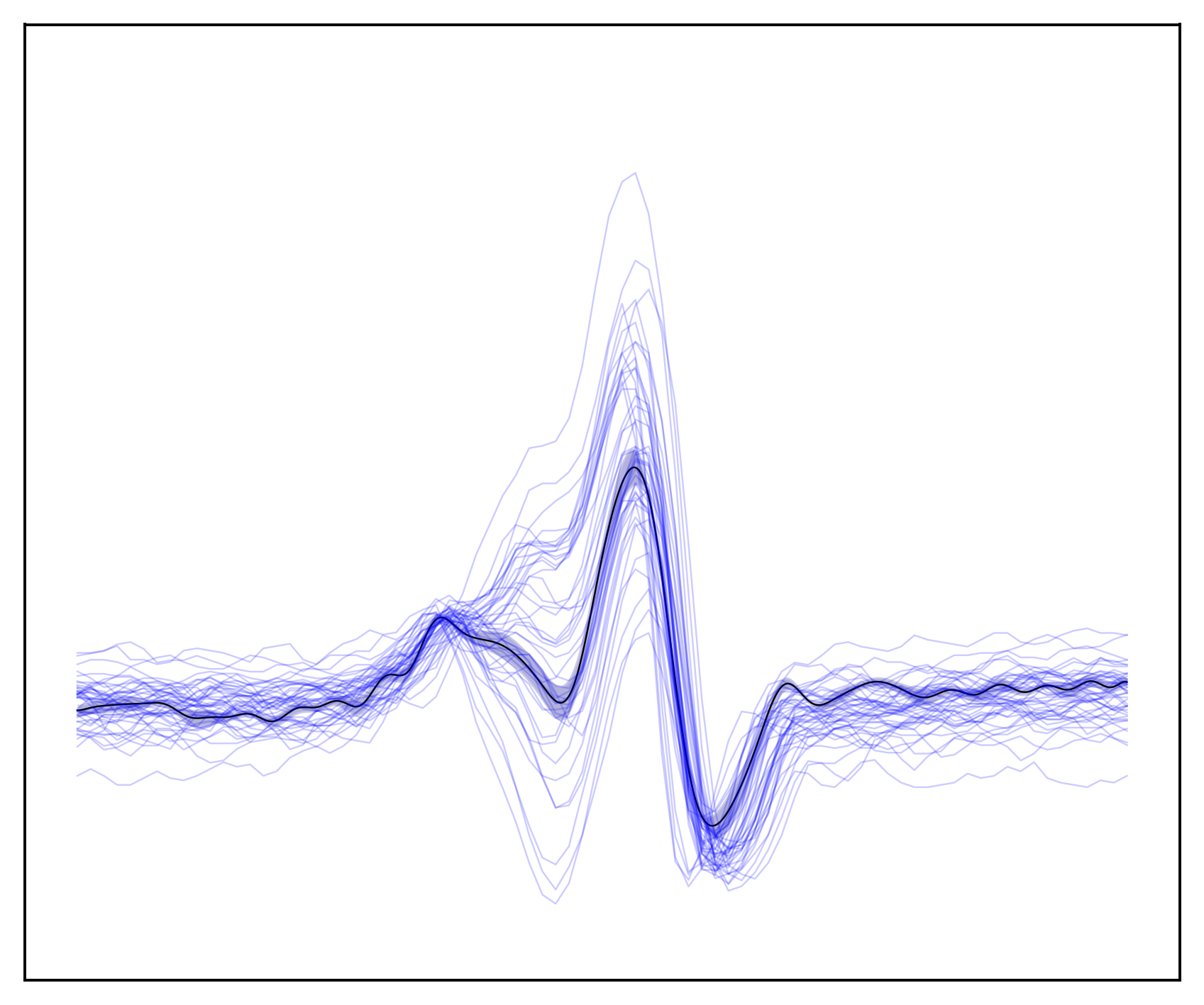}
    \includegraphics[width=0.35\columnwidth]{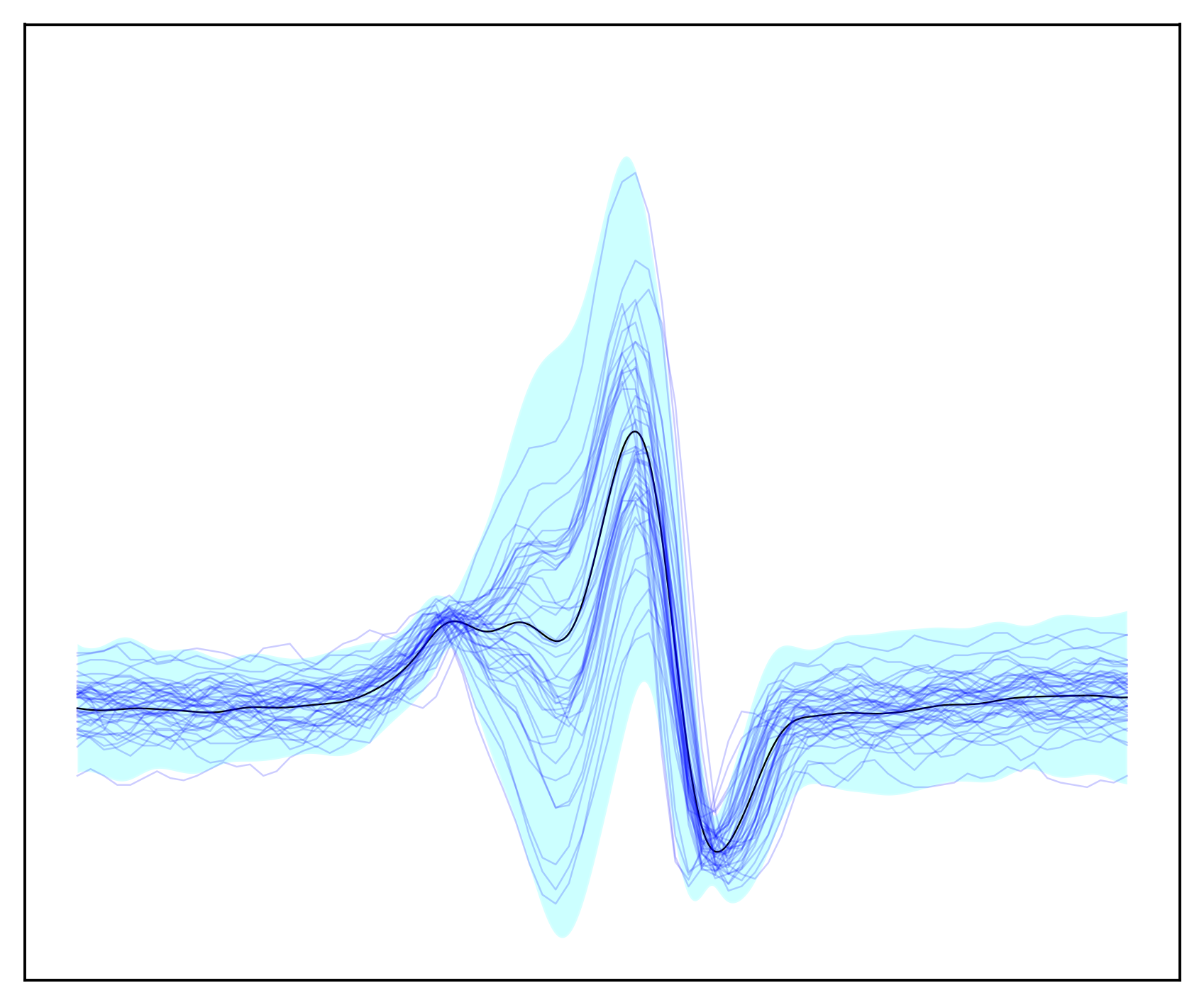}
    \caption{Model enhancing stability. The same 50 heartbeats as in Figure~\ref{fig:plasticity_model} are represented, from record 114 of the MIT-BIH Arrhythmia Database. The resulting GP $f(t)$ (left) and $x(t)$ (right), after evolving along with these sequences, were sampled over a dense time axis. Kernel hyperparameters for $k_{\theta}(t,t')$ are $\sigma^{2}_{f} = 16.0^{2}$, $\ell = 2.7$, $\sigma^{2}_{n} = 5.0^{2}$. LDS priors: $S_{\omega} = 4.0^{2}I, S_{\epsilon} = 15.0^{2}I$. For clarity purposes, no warping function has been used. Again, the model embraces the totality of the heartbeats in a single cluster.}
    \label{fig:stability_model}
    \Description[Stable GP model for ECG beats]{
    Two-panel ECG plot showing the same fifty normal heartbeat segments under a stable configuration. The posterior mean represents an average beat morphology, and the uncertainty bands become wider enough to describe variability around this representative.
    }
\end{figure}

The covariance priors $S_\omega$ and $S_\epsilon$, and particularly, the magnitude of disparity between them, play a key role in managing the stability-plasticity balance of the resulting model, namely the preference for enhancing the ability to adapt the current clusters to new data (plasticity) versus the preference for enhancing the capacity to retain previously learnt clusters (stability). As $S_{\omega} > S_{\epsilon}$, the model tends to explain a succession of sequences by making evolve each cluster according to the respective linear dynamics, hence enhancing plasticity (see Figure~\ref{fig:plasticity_model}). Conversely, as $S_{\omega} < S_{\epsilon}$, the model tends to explain the same succession of sequences by considering their evolution as the result of some sort of variability around an average sequence (see Figure~\ref{fig:stability_model}). So far, we are assuming that all the clusters raised by the model share the same covariance priors, and therefore a single stability-plasticity criterion is enforced. But the availability of previous knowledge about the problem can enable us to specify different covariance priors for different categories, in a semisupervised fashion.

We consistently set $W_A=S_\omega$ and $W_C=S_\epsilon$, and the inverse-Wishart portions are given $\nu_0=p+2$ and $\varsigma_0=q+2$ degrees of freedom, respectively.

Each LDS models a linear evolution of a GP with prior $f^{k}(t) \sim \mathcal{GP}(m_{\theta ^{k}}(t),k_{\theta ^{k}}(t,t'))$, where mean $m_{\theta ^{k}}(t)$ is usually chosen to be $\mathbf{0}$ due to symmetry. A reasonable choice for the covariance function $k_{\theta ^{k}}(t,t')$ commonly used in the GP literature when no prior information is available is the squared exponential kernel, although any kernel could be used within our method~\citep{rasmussen_gaussian_2006}. The squared exponential kernel has the form $k_{\theta ^{k}}(t, t') = \sigma^{2}_{f} \exp{(-\frac{1}{2 \ell^{2}}(t-t')^{2}}) + \sigma^{2}_{n}\delta_{t,t'}$ with hyperparameters $\theta=(\sigma_{f}, \ell, \sigma_{n})$. The length-scale $\ell$ controls the temporal variation necessary for the sequence values to appreciably change; the signal variance $\sigma^2_{f}$ controls the variation of function values from their mean; and the noise variance $\sigma^2_{n}$ controls the noise which is expected to be present in the observations. These hyperparameters are determined by maximum likelihood from the dataset. 

Each warping function models the misalignment between the observations and the predicted pseudo-observations, with prior $g^{k}(t) \sim  \mathcal{GP}(m_{\vartheta ^k}(t),k_{\vartheta ^k}(t,t'))$. The mean $m_{\vartheta ^k}(t)$ has been chosen to be $\mathbf{0}$ around the linear unit function. A squared exponential kernel has also been selected for the covariance function $k_{\vartheta ^k}(t,t')$ since it yields smoothing properties. Reducing the length-scale hyperparameter or using a non-smooth kernel, like the Matérn kernel, can improve alignment functions and model time alignments with skipping events.

\begin{figure}[!t]
    \centering
    \includegraphics[width=1.0\textwidth]{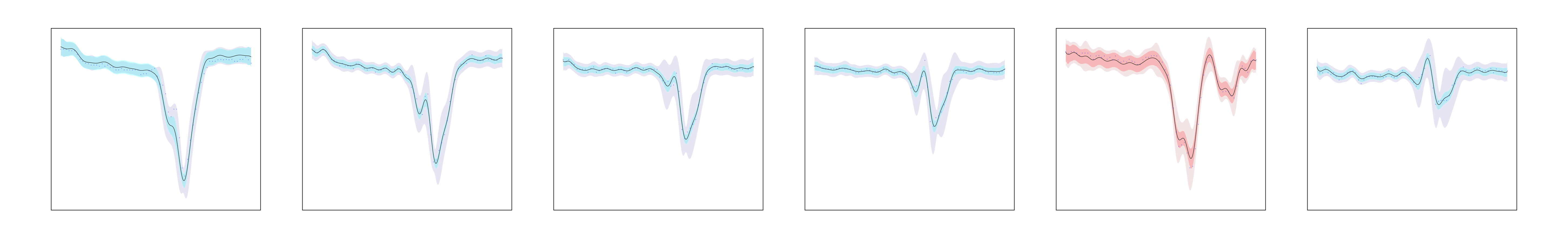}
    \caption{The GP model for each one of the beats depicted in Figure~\ref{fig:ecg_e1302} is shown. Our method correctly includes all the beats in the same evolving cluster, except for the fifth one, which gives rise to a new cluster.}
    \label{fig:evolution_models_e1302}
    \Description[Evolving ECG cluster models]{
    Grid of Gaussian-process models associated with the ECG beats shown earlier. Beats previously identified as normal are represented by the same evolving morphology even considering the amplitude variations, while the disruptive ventricular beat is assigned to a separate cluster.
}
\end{figure}

\section{Applications}
\label{sec:applications}
\subsection{ECG heartbeat clustering}

Figure~\ref{fig:evolution_models_e1302} illustrates the expected behaviour of the proposed model: a cluster initially generated from a Normal beat evolves to absorb the gradual morphological changes of the same conduction pattern, whereas a new cluster is created when a genuinely different morphology, such as a Ventricular beat, appears.

A quantitative evaluation was performed on the MIT-BIH Arrhythmia Database~\citep{goldberger_2000}, a standard reference for ECG arrhythmia analysis~\citep{moody2001}. The database contains 48 excerpts of two-channel ambulatory ECG recordings from 47 subjects, sampled at 360 Hz for 30 minutes. Each beat was annotated by at least two expert cardiologists using a 17-label annotation set.

Following common ECG morphology-clustering practice, we focus on the QRS complex, which mainly represents ventricular depolarisation. Therefore, atrial activity and rhythm information are intentionally excluded from the representation. This design choice makes the evaluation stricter: beats with different atrial origins but similar QRS morphology may be assigned to the same cluster, even if their clinical labels differ.

We compare the number of clusters generated by the proposed method with the context-based adaptive method of~\citep{castro2015}, excluding rhythm information. That method obtains a global purity of 97.15\%, while generating 629 clusters across the full database. The proposed single-output offline method obtains a global purity of 97.12\% with 335 clusters, and the online method obtains 97.31\% with 539 clusters. The multi-output offline version, which uses both ECG leads through the SNR-weighted likelihood, improves the purity to 97.53\% while generating 342 clusters. Detailed comparisons with additional metrics and deep clustering baselines are reported in Appendix~\ref{app:mitbih_results}.

Table~\ref{tab:clusters_per_record} reports the number of clusters generated for each MIT-BIH record. The proposed method substantially reduces the number of clusters with respect to~\citep{castro2015}: 45.6\% for the multi-output offline method, 46.7\% for the single-output offline method, and 14.3\% for the online method. This reduction is the main indication of the parsimony induced by the model: gradual morphological variation tends to be absorbed by the dynamical evolution of an existing cluster instead of generating redundant templates.

\begin{table}[ht] 
\setlength{\tabcolsep}{3pt}
\renewcommand{\arraystretch}{1.1}
\caption{Number of clusters per record in the MIT-BIH Arrhythmia Database.} 
\label{tab:clusters_per_record} 
\centering{
\begin{tabular}{c c c c c | c c c c c} 
\hline \hline
Record &~\citep{castro2015} & $N_{MO}$ & $N_{OFF}$ & $N_{ON}$ &
Record &~\citep{castro2015} & $N_{MO}$ & $N_{OFF}$ & $N_{ON}$\\ \hline
100 & 4  & 2  & 3  & 2  & 201 & 15 & 10 & 6  & 10 \\
101 & 4  & 2  & 3  & 2  & 202 & 9  & 7  & 8  & 6  \\
102 & 10 & 10 & 9  & 8  & 203 & 33 & 7  & 12 & 38 \\
103 & 10 & 4  & 5  & 2  & 205 & 14 & 10 & 7  & 9  \\
104 & 16 & 15 & 10 & 27 & 207 & 61 & 9  & 15 & 36 \\
105 & 10 & 4  & 10 & 50 & 208 & 28 & 17 & 11 & 30 \\
106 & 27 & 4  & 8  & 11 & 209 & 10 & 6  & 9  & 6  \\
107 & 11 & 6  & 3  & 11 & 210 & 27 & 9  & 9  & 11 \\
108 & 22 & 2  & 8  & 9  & 212 & 5  & 3  & 4  & 5  \\
109 & 13 & 8  & 9  & 11 & 213 & 17 & 15 & 12 & 23 \\
111 & 8  & 2  & 5  & 5  & 214 & 21 & 12 & 12 & 18 \\
112 & 4  & 3  & 2  & 2  & 215 & 16 & 16 & 7  & 10 \\
113 & 5  & 3  & 3  & 2  & 217 & 28 & 11 & 10 & 22 \\
114 & 8  & 4  & 4  & 5  & 219 & 14 & 12 & 10 & 10 \\
115 & 11 & 3  & 4  & 3  & 220 & 2  & 3  & 7  & 3  \\
116 & 10 & 7  & 7  & 9  & 221 & 14 & 7  & 9  & 6  \\
117 & 4  & 4  & 3  & 3  & 222 & 8  & 4  & 3  & 3  \\
118 & 3  & 5  & 6  & 12 & 223 & 23 & 11 & 10 & 20 \\
119 & 6  & 3  & 4  & 5  & 228 & 14 & 14 & 10 & 16 \\
121 & 5  & 4  & 5  & 6  & 230 & 3  & 4  & 3  & 6  \\
122 & 1  & 1  & 2  & 2  & 231 & 5  & 4  & 7  & 6  \\
123 & 3  & 3  & 3  & 2  & 232 & 4  & 2  & 3  & 3  \\
124 & 14 & 9  & 7  & 7  & 233 & 24 & 21 & 10 & 20 \\
200 & 20 & 16 & 13 & 22 & 234 & 5  & 4  & 5  & 3  \\ \hline
&&&& TOTAL &  & 629 & 342 & 335 & 539 \\ \hline \hline
\\
\end{tabular}}
\centering

{Column $N_{MO}$ stands for the number of clusters generated by the multi-output offline strategy using both leads with SNR weighting; $N_{OFF}$ stands for the number of clusters generated by the single-output offline strategy; and $N_{ON}$ stands for the number of clusters generated by the single-output online strategy.}
\end{table}

\begin{figure}
    \centering
    \includegraphics[width=0.85\linewidth]{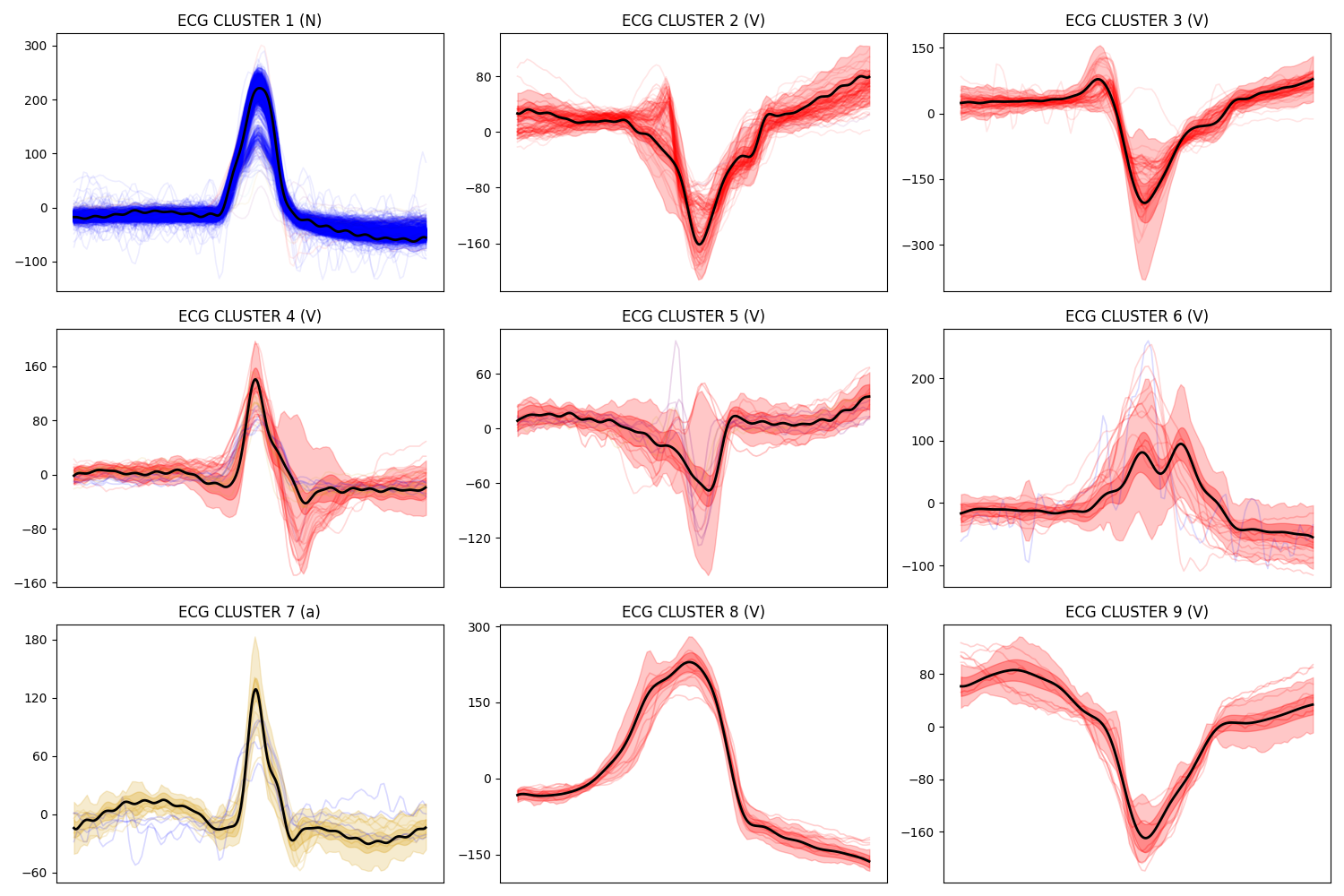}
    \caption{Clusters identified from record 210 of the MIT-BIH Database. The black line denotes the posterior mean, the pink region represents the 95\% confidence interval for the latent process $f(\cdot)$, and the red region corresponds to the emission process $x(\cdot)$. Cluster 1 captures nearly all normal heartbeats despite gradual morphological evolution. Cluster 7 represents an aberrated atrial premature morphology, while the remaining clusters correspond to multiform premature ventricular contractions.}
    \label{fig:enter-label}
    \Description[ECG clusters from MIT-BIH record 210]{
    Grid of ECG morphology clusters inferred from MIT-BIH record 210. Each panel overlays multiple aligned beat traces with a posterior mean curve and uncertainty bands. One large cluster contains most normal beats, while smaller clusters contain ventricular or aberrated morphologies.
    }
\end{figure}

Figure~\ref{fig:enter-label} illustrates the clusters obtained in an offline fashion for record 210. The dominant cluster captures most normal beats despite their gradual morphological evolution. Other clusters correspond to multiform premature ventricular contractions, reflecting distinct ventricular activation sites, and to an aberrated atrial premature morphology.

Most of the beats in the database are annotated as normal (68.3\%), and most normal beats are assigned to clusters in which they represent the majority: 99.7\% for both offline strategies and 99.8\% for the online strategy. The most common type of error is that abnormal beats are assigned to clusters dominated by normal beats, accounting for 56.3\% of the errors in the multi-output offline method, 52.5\% in the single-output offline method, and 54.1\% in the online method. Beats with supraventricular or nodal activation are the main source of errors, since their discrimination often requires atrial activity or rhythm information, both intentionally excluded from the representation. They account for 49.5\% of the errors in the single-output offline method and 52.9\% in the online method. Escape beats from record 222 account for 15.5\% of the errors in the multi-output offline method, 6.7\% in the single-output offline method, and 7.2\% in the online method. Recordings performed on patients with pacemakers, particularly records 102, 104 and 217, contain fusion beats resulting from the concurrent activation of the same ventricular region by the intrinsic beat and the pacemaker impulse. These beats account for 8.7\% of the errors in the multi-output offline method, 7.9\% in the single-output offline method, and 9.6\% in the online method. In this sense, it can be argued that the smoothing properties of the squared-exponential kernel yield blurred representations of pacing pulses, and hence they may be added as slight variations of previous models. Other kernels could better manage these signal discontinuities. For a description of the hyperparameter choices see Appendix~\ref{app:hyperparams_mitbih}.

\subsection{Breathing estimation}
The present approach allows us to pose new questions on the dynamics underlying the available observations. Let us focus, for instance, on the sequence of warping functions used to perform beat clustering. These warping functions capture the specific misalignment of each observed beat with respect to the predicted one, and we can wonder about whether they contain any physiological information. Indeed, the observation of some regularity in the sequence of warping functions may lead us to conjecture that the misalignments of a sequence of beats with the same morphology result to a large extent from the breathing mechanism, since the electrical activity of the heart has to pass through the lungs before it is measured by the electrodes placed on the chest. The inspiration and expiration cycle changes the volume of air, with strong dielectric properties, in the conduction pathway to the electrodes, cyclically delaying the observation of the different constituents of the heartbeat. As a consequence, the sequence of warping functions would show a correlation with the breathing cycle, allowing its estimation.

To test this hypothesis, a supervised reconstruction experiment has been devised on two public databases, the Fantasia Database~\citep{Iyengar1996}, consisting of ECG and respiration recordings from 40 healthy subjects in sinus rhythm during a resting state; and the Apnea-ECG Database~\citep{Penzel2000}, including 8 ECG recordings accompanied by different respiration signals and apnea annotations. We consider two time series, the respiration signal denoted by $\{(t_{s},r_{s})\}_{s=1}^S$ and the sequence of different warping segments obtained from the ECG signal, denoted by $\{(\mathbf{t}_{n},\mathbf{t}_{n}^{w})\}_{n=1}^N$. Let us note that this last sequence is not defined for the whole temporal range of the respiration signal, as the warping function is just computed for each QRS temporal interval. The sequence of warping segments is obtained by Equation (\ref{eq:maximizing_g}). Thus, an initial GP model $\mathbf{f_1}$ is computed from the first normal beat $(\mathbf{t}_1,\mathbf{y}_{1})$ of the ECG recording and then an alignment function is computed for each one of the following beats. To emphasise the alignment differences between consecutive beats, the initial model has not been evolved with new beats, hence preventing the LDS from absorbing some of the alignment variability.

\begin{figure}
    \centering
    \includegraphics[width=\textwidth]{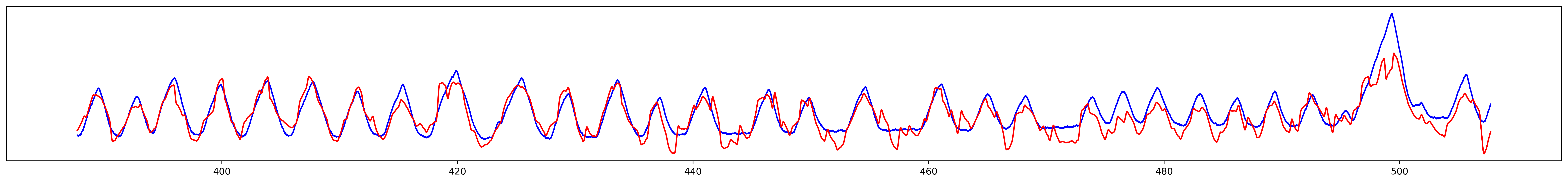}
    \caption{Respiration signal (in blue) and prediction (in red) for record f1y01, [6:30.000,8:30.000] of the Fantasia Database is shown. Training set involves a total of 500 beats from [0:00.000,6:30.000].}
    \label{fig:breathing_test_fy01}
    \Description[Respiration reconstruction for Fantasia record f1y01]{
    Time-series plot comparing the recorded respiration signal with the respiration signal reconstructed from ECG-derived warping functions for Fantasia record f1y01. The two curves show similar oscillatory behaviour over the displayed interval.
}

    \includegraphics[width=\textwidth]{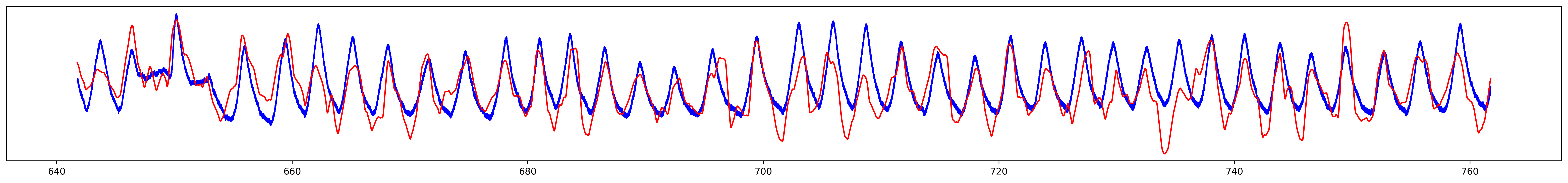}
    \caption{Respiration signal (in blue) and prediction (in red) for record f1o03, [10:42.000,12:42.000] of the Fantasia Database is shown. The training set involves a total of 500 beats from the beginning of the recording.}
    \label{fig:breathing_test_f1o03}
    \Description[Respiration reconstruction for Fantasia record f1o03]{
    Time-series plot comparing the recorded respiration signal with the respiration signal reconstructed from ECG-derived warping functions for Fantasia record f1o03. The reconstructed curve follows the dominant respiratory rhythm, although some local amplitude and phase differences remain.
}

    \includegraphics[width=\textwidth]{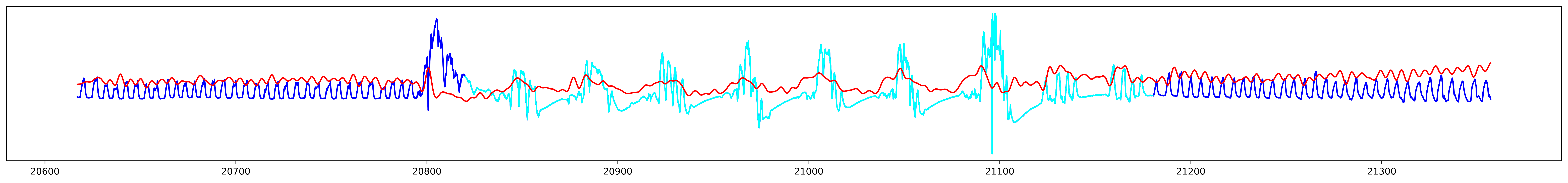}
    \caption{Respiration signal (in blue) and prediction (in red) for record a03 (seconds 20640 to 21360), [5:44:00.000,5:56:00.000] of Apnea-ECG Database is shown. An apnea episode (in cyan) is annotated in the record. Prediction lacks accuracy due to a low sampling rate (100Hz), but the period of the respiration signal can be correctly estimated. The training set involves a total of 1500 beats from the beginning of the recording, including the occurrence of apnea episodes.}
    \label{fig:breathing_test_a03}
    \Description[Respiration reconstruction during apnea]{
    Time-series plot comparing recorded and reconstructed respiration for Apnea-ECG record a03. The figure includes an annotated apnea interval, during which the respiratory pattern changes; the reconstruction mainly captures the respiratory periodicity with a visible change on the apnea interval.
}
\end{figure}

We focus on those respiration signal segments $\{(\mathbf{t}_n,\mathbf{r}_{n})\}_{n=1}^N$ for which there is a warping segment in the same time interval $(\mathbf{t}_n,\mathbf{r}_n)=\{(t_s,r_s):t_s\in\mathbf{t}_n\}$, so trying to find a relation between them can be approached as a usual supervised estimation problem. Assuming linearity as a naive approach, estimation can be posed as a least squares fitting problem, where $\mathbf{r}_{n} = M_{w} \mathbf{t}_{n}^{w}$. For testing the robustness of this approach, we estimate the matrix $M_{w}$ from a training set of both respiration and warping segments at the beginning of the recording and reconstruct the respiration signal from then on with subsequent smoothing. Estimation results from some respiration recordings are shown in Figures~\ref{fig:breathing_test_fy01},~\ref{fig:breathing_test_f1o03} and~\ref{fig:breathing_test_a03}.

\subsection{Wave-spectra}

Understanding wave conditions is essential for applications ranging from maritime navigation and offshore engineering to coastal flood prediction and climate modelling~\citep{Holthuijsen_2007}. Traditionally, deployed buoys measure time series of water elevation at a point and wave conditions are summarised by bulk scalar parameters such as significant wave height, peak period, and mean direction. While these metrics are convenient for real-time reporting and climatological summaries, they represent sparse projections of the full wave field onto low-dimensional statistics~\citep{Hamilton_2010}. In particular, a sea state may contain a narrow swell peak, a locally generated wind-sea component, or several simultaneous systems arriving from different directions. These properties can be distinguished using the directional wave-spectra representation~\citep{Pierson_1964}: the frequency axis describes the spectral morphology, whereas the directional bins describe the origin of the phenomena. We therefore use this example to test the multi-output version of the model, where the potential regimes evolve over time and the number of sea-states is unknown a priori.

Let $S_n(\nu,\theta)$ denote the directional spectrum observed at time $n$, where $\nu$ is frequency and $\theta$ is direction. We represent each spectrum as a multi-output observation by treating each directional bin $\theta_\ell$ as one output channel,
\begin{equation}
    \mathbf{y}^{(\ell)}_n
    =
    S_n(\nu,\theta_\ell),
    \qquad
    \ell=1,\ldots,L.
\end{equation}
All directional outputs share the same latent regime assignment $s_n$, while their contribution to the likelihood can be weighted according to the energy contained in each direction. Specifically, we use
\begin{equation}
w^{(\ell)}_n
=
\frac{
\sum_j S_n(\nu_j,\theta_\ell)
}{
\sum_{\ell'=1}^{L}\sum_j S_n(\nu_j,\theta_{\ell'})
},
\end{equation}
so that directions carrying most of the wave energy contribute more strongly to the cluster assignment. This avoids modelling the full frequency--direction covariance explicitly, while still coupling all directional slices through a shared discrete state sequence. 

\begin{figure}
    \centering
    \includegraphics[width=0.95\textwidth]{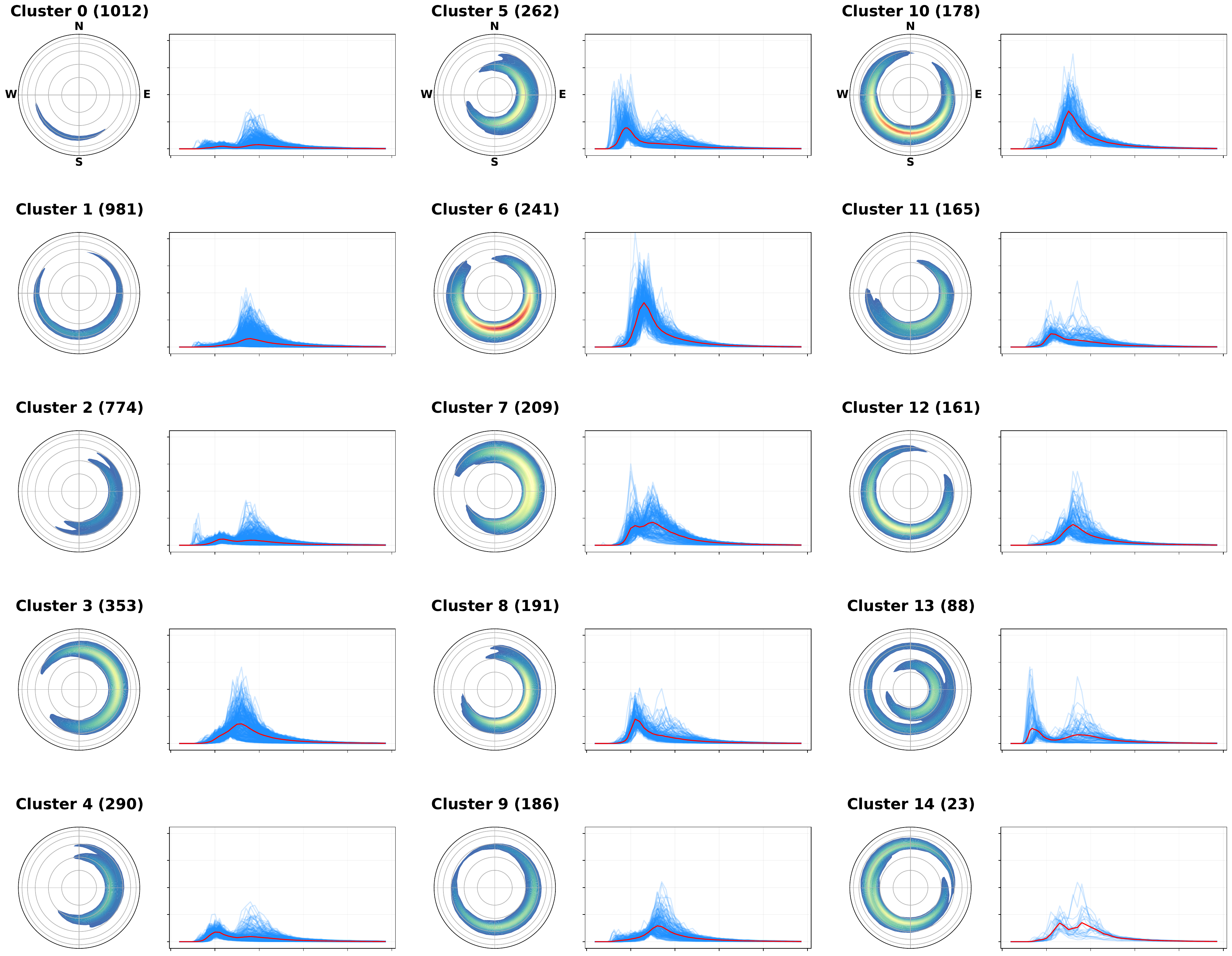}
    \caption{Directional wave-spectra regimes inferred by the multi-output HDP-GPC model at NDBC station 41013. For each identified regime, the number of wave spectra is included in parentheses. The polar plot shows the mean directional distribution of spectral energy measured in m$^{2}$/Hz, the right column shows the direction-integrated frequency spectra of the cluster and the GP representative.}
    \label{fig:wave_spectra_clusters}
    \Description[Directional wave-spectrum clusters]{
    Summary of inferred directional wave-spectrum regimes. Each cluster is represented by a polar plot of mean directional energy and a frequency-spectrum plot with the assigned spectra and Gaussian-process representative, showing the dominant direction and spectral morphology of each sea-state cluster.
}
\end{figure}

For the experiment shown in Fig.~\ref{fig:wave_spectra_clusters}, HDP-GPC was applied to directional wave spectra from NDBC station 41013, located at Frying Pan Shoals, off the coast of Cape Fear, North Carolina, during 2017 and 2018. The original NDBC files provide the one-dimensional spectral density over 47 frequency bands, approximately from 0.02 Hz to 0.485 Hz, together with directional parameters that allow the reconstruction of the two-dimensional spectrum. The directional spectrum was reconstructed over 36 angular bins, and the analysis was restricted to the informative frequency range below 0.3 Hz. Following~\citep{Hamilton_2010}, spectra with significant wave height outside the interval 0.5--2.5 m were discarded, and the data were averaged every three hours to reduce computational load while preserving the temporal evolution of the sea state.


The model inferred $K=15$ regimes from a total of 5114 spectra. According to the partition criteria introduced in~\citep{Portilla_2009}, the inferred clusters exhibit combinations of swell, wind sea and transition sea. The first three clusters contain the 54\% of the observations. Cluster 0 is characterised by a very frequent superposition of groundswell and a wind swell induced by south-west winds, both with low energy. The spectrum of cluster 1 is dominated by a single peak around 0.15 Hz and a broad southern directional spread, which makes it a moderate-energy intermediate regime between locally forced wind sea and a freely propagating swell, a typical sea-state from summer or spring cycles. Cluster 2 structure is close to Cluster 1, considering the wind component, but also includes a low-frequency swell pattern, especially compatible with a residual incoming or decreasing swell. 

Clusters 3 to 12 represent more clearly separated morphologies. Cluster 3 is a high-energy wave regime with a wide direction spread oriented north-easterly, and its frequency characterises a mature and stable sea state recurrent in winter or autumn. Cluster 4 is an active transition wind-sea pattern, with a well-defined peak close to 0.1 Hz and a dominant western direction indicating potential local wind conditions. Cluster 5 is a fully developed swell-dominated system propagating from the eastern sector, potentially associated with sporadic storms far from the coastline, mixed with a remanent local-wind component. Cluster 6 presents a coherent, persistent high peak in the lower-frequency range, compatible with an old wind sea state typical of winter and autumn climate patterns. Cluster 7 represents a sea state with widespread frequencies indicating transitional sea conditions with north-east components. In cluster 8, a low-frequency peak dominates, but the directional spectrum is not as clearly defined as cluster 5, which indicates a well-developed swell overlapping with more systems. Its occurrence is cyclical, happening four times in 2017 and five times in 2018, mainly in spring and autumn. Cluster 9 is a weak energy and smooth spectrum in frequency and direction, indicating a background regime with a wind component from the south-east. Cluster 10 is a strong-wind storm from the south-west, a repeatable pattern for late winter-early spring, determined by a high peak in a well-determined frequency but a wide directional spread. Cluster 11 is a mixed sea state determined by a low-energy swell component coming from the west and potentially the same wind component present on cluster 10 coming from the south-west. Cluster 12 represents a transition sea state between a background state and the wind-generated cluster 10, working as an announcement of stronger wind conditions.

The remaining clusters contain fewer observations and correspond to more specific spectral configurations. Cluster 13 combines a strongly developed low-frequency swell component from the eastern sector with a second wind-sea component that is directionally dispersed. Finally, cluster 14 represents the coexistence of a very weak swell component and a wind-sea component directed towards the south-west. Its clean structure and low number of observations indicate a specific and short-lived situation, but with a strong repeatability during both years analysed, as it only happens twice a year in early spring and late winter. For a detailed hyperparameter choices see Appendix~\ref{app:wave_hyperparameters}.

This experiment illustrates the usefulness of a nonparametric clustering formulation in this setting. Compared with CLARA clustering proposed by Hamilton~\citep{hamilton2010}, where silhouette elbow selected $K=31$ clusters in this experiment, HDP-GPC inferred a more compact set due to explicitly modelling temporal evolution. Moreover, by using directional spectra rather than only frequency- or direction-marginal descriptors, the clustering retains information about both the frequency-domain morphology and the directional distribution of the wave energy. The partitioning and labelling procedure then provides an additional physical and temporal description of the inferred clusters. 

\subsection{Pedestrian trajectory clustering}

We used the spatiotemporal trajectory database of pedestrian mobility at train stations introduced in~\citep{Apeltauer_2024}. The dataset contains anonymous pedestrian trajectories collected at the train stations of Havl\'{\i}{\v c}k\r{u}v Brod and Pardubice, Czech Republic, from September 2022 to August 2023. Pedestrian trajectories encode recurring movement patterns imposed by the surrounding structures, and separating those trajectories into route classes allows different movement patterns to be modelled individually, supporting downstream applications such as design of station layouts, crowd management strategies and evacuation procedures~\citep{Xue_2021}. The data were acquired using 11 cameras and transformed into tabular trajectories in a local metric coordinate system. For simplicity, we applied the proposed method to camera PB-CAM6 on the day of 9th of January 2023.

In contrast to ECG beats or wave spectra, dynamic clustering is difficult to justify for individual pedestrian trajectories because the samples do not form a single temporal chain: many pedestrians move concurrently, trajectories overlap in time, and ordering them by start time, end time, or midpoint can produce different and artificial sequences. As an example, a long trajectory may contain several shorter trajectories inside its duration, so a transition from one sample to the next may reflect only the chosen sorting rule rather than a real evolution of crowd flow. A more robust approach is to cluster trajectories statically by shape, and then study dynamics afterwards by studying the developing times. Each record contains a trajectory identifier, relative time, date, and the spatial coordinates $(x,z)$.

Each pedestrian path is represented as a two-output functional observation,
\begin{equation}
    \mathbf{y}_n(\tau)
    =
    \big(x_n(\tau),z_n(\tau)\big),
    \qquad
    \tau \in [0,1],
\end{equation}
where $\tau$ denotes normalised trajectory progress. We resampled trajectories onto a common support and clustered them using a static version of the HDP-GPC model. This corresponds to collapsing all transition rows to the global HDP weights,
\begin{equation}
    \boldsymbol{\pi}^{(j)} = \boldsymbol{\beta},
    \qquad
    s_n \sim \boldsymbol{\beta},
\end{equation}
so that assignments are exchangeable rather than Markovian. In addition, the dynamical matrices are fixed to the identity, $\mathbf{A}^{(k)}=\mathbf{I}$, $\mathbf{C}^{(k)}=\mathbf{I}$, which removes the temporal evolution of each cluster representative. Hence, observations are emitted directly from the representative GP associated with the selected route family,
\begin{align}
\mathbf{f}^{(k)}(\cdot)
    &\sim
    \mathcal{GP}
    \left(
        m_{\theta^{(k)}}(\cdot),
        k_{\theta^{(k)}}(\cdot,\cdot)
    \right)\\
    \mathbf{y}_n(\tau)
    &=
    \mathbf{f}^{(s_n)}(\tau)
    +
    \boldsymbol{\epsilon}_n(\tau).
\end{align}
Thus, the model keeps the nonparametric GP-based representation of route families, but avoids imposing artificial temporal dependence between consecutive trajectories in the file.

\begin{figure}
    \centering
    \includegraphics[width=0.6\textwidth]{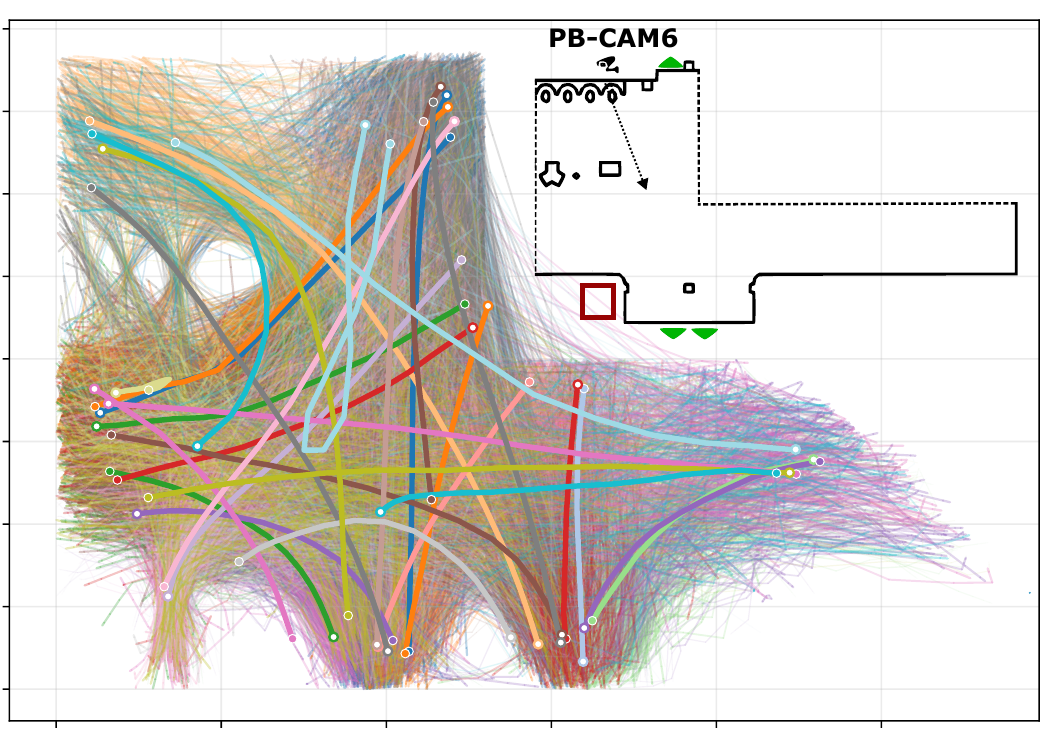}
    \caption{Overview of pedestrian trajectory clustering for PB-CAM6, the 9th of January 2023, using the static HDP-GPC model. Coloured curves show the observed pedestrian trajectories, and bold curves show the GP representatives of the inferred route families. The walking direction is illustrated by an empty point indicating the initial position and a filled one indicating the ending position. The top-right diagram shows a map of the tracked zone from PB-CAM6, with green arrows indicating exits and entrances, the red box indicating the position of a shop, solid lines indicating obstacles and dashed lines indicating the limits of the camera-tracked zone.}
    \label{fig:pedestrian_clusters}
    \Description[Pedestrian trajectory clustering overview]{
    Overview of pedestrian trajectories recorded by camera PB-CAM6. Many coloured paths are overlaid with thicker Gaussian-process route representatives. A small layout diagram marks entrances, exits, obstacles, a shop area, and the camera-tracked region.
}
\end{figure}

Figure~\ref{fig:pedestrian_clusters} shows the inferred movement patterns for the trajectories recorded on the 9th of January 2023. The figure overlays all trajectories from that day together with the GP representatives of the inferred route families. A complete visualisation of the individual clusters and a detailed selection of hyperparameters is reported in Appendix~\ref{app:pedestrian_clusters}. Note that the direction was highly influential on the results, providing two clusters for similar routes but with opposite directions. The largest clusters correspond to the dominant station corridors, whereas smaller clusters capture less frequent yet coherent movement patterns.

This experiment shows that HDP-GPC can be used as a hierarchical mixture model for Bayesian nonparametric clustering of time series in non-dynamic settings. In contrast to PoPPL~\citep{Xue_2021}, where route classes are built from source and destination regions, potentially missing differences between trajectories that share endpoints but follow different paths, the present proposal does not reduce pedestrian motion, but each path is modelled, allowing the full route to contribute to the clustering. Similarly, unlike deep representation learning~\citep{Yao_2017}, which relies on a latent embedding after a feature-engineering stage, the HDP--GPC keeps the cluster representation directly interpretable in the original coordinate system.


\section{Efficient approximations}
Optimizing the hyperparameters of both the initial GP of each cluster and the warping GP computed for each new segment involves a computational complexity driven by the inversion of the covariance matrices \( K^{\theta^{(s_{n})}}_{\mathbf{t}_1,\mathbf{t}_1} \) and \( K^{\vartheta^{(s_{n})}}_{\mathbf{t}_n,\mathbf{t}_n} \), respectively, each scaling with \(\mathcal{O}(q^3)\). To mitigate this, several efficient approximations have been proposed, including methods from~\citep{quinonero-candela05, rasmussen_gaussian_2006, csato2000}. A notable approach by Seeger~\citep{Seeger2005phd}, stated in Equation \eqref{eq:seeger_approach}, reduces complexity to \(\mathcal{O}(p^3)\) by using a reduced number \( p < q \) of inducing points.

The computation of the warp function remains the most demanding component of the proposed framework. In fact, the online strategy attained real-time performance for heartbeat clustering when warping computation was excluded, since the message-passing operations involved in variational inference are efficiently handled using a forward-backwards approach. Optimising the hyperparameters of the warping GP is currently solved by maximum a posteriori (MAP) estimation; however, convergence can be slow and computationally expensive. A promising approach is introduced in~\citep{mikheeva2022}, where the alignment is achieved using monotonic processes defined as solutions to Ordinary Differential Equations (ODEs) with uncertain drift functions. These drift functions are modelled as GPs, providing smooth and monotonic warps while allowing for uncertainty quantification. To sample the monotonic processes, an efficient path-wise sampling technique is used, reducing complexity to \(\mathcal{O}(q)\). The empirical effect of these computational choices is quantified in Appendix~\ref{app:mitbih_results}, where runtime and memory are compared across warped, unwarped, single-output and multi-output configurations.

The proposed formulation has three main limitations. First, the current
multi-output extension uses a weighted composite likelihood and does not
explicitly model cross-output covariance. Second, temporal alignment is
estimated by MAP, so uncertainty over the warping functions is not propagated
to the remaining latent variables. Third, the computational cost grows with
the number of active clusters and becomes dominated by the repeated warp
optimisation step when alignment is enabled. These limitations motivate the
fully coupled multi-output and more efficient monotone-warp extensions
discussed below.

\section{Related work}
\label{sec:related_work}
Probabilistic approaches to clustering time-series data have been widely explored, particularly through mixture models~\citep{smyth1996, biernacki2000, zhong2003, bach2004}. These methods assume that the data is generated by a mixture of distributions with predefined forms, such as Gaussian, HMMs, or graphical models. With the aim of reducing the strong modelling assumptions, a nonparametric approach is proposed by~\citep{kahlegi2016} in exchange for assuming stationary ergodicity over the data distribution. While effective for static data, these methods often fail to account for the evolving nature of time series, where distributional shifts can reduce precision, and where a fixed number of clusters is not flexible enough to deal with the emergence of new clusters. Trying to reduce this variability between observations, alignment based on GPs is developed in~\citep{kazlauskaite_2019}, resulting in better clustering within the latent space. Also, works such as~\citep{chiu2022, kolter2007} address this variability by detecting and reacting to concept drift. In contrast, the present approach integrates this dynamic transformation directly into the model, enabling clusters to adapt independently while parameterising the evolving distributions over time.

Bayesian nonparametrics have provided a natural solution to overcome the limitation of fixed cardinality for the state space. In~\citep{qi2007}, the authors propose an HDP-HMM model for clustering fixed-length segments of a time series, with applications to music analysis. In~\citep{chakraborty2023}, an infinite hidden Markov model (iHMM) on a
state space of multi-dimensional vector fields described by GPs is proposed to extract traffic patterns from multi-vehicle trajectories. In~\citep{fox_2011}, HDP-SLDS has proven effective for modelling temporal dynamics in time-series data, with applications to segmentation of dancing honey bee trajectories and learning changes in the volatility of the IBOVESPA stock exchange. A sticky version of the original model increases the expected probability of self-transition, thus avoiding unrealistic fast changes between linear regimes caused by the intrinsic variability of raw data. The present proposal avoids this problem through pattern abstraction on the time-series data, and therefore, the segmentation is performed using time-series segments as samples, increasing the stability of the required dynamics. 
    
Procedures for efficient learning of nonlinear state-space models and nonlinear mappings from latent states to observations have been proposed in~\citep{Damianou_2011, frigola14}, by placing proper GP priors. These methods exhibit good performance in applications such as reconstructing human motion capture data and high-dimensional video sequences, but they need more complex non-conjugate inference and learning algorithms. Notably, in~\citep{alvarez09}, the latent force model is introduced as a combination of mechanistic modelling principles with non-parametric data-driven components, using GPs as non-parametric models for unknown input functions in physical models, which are formulated as ordinary differential equations (ODEs). Applications in modelling human motion capture data, describing the time-dependent expression levels of genes, and predicting heavy metal concentrations in geostatistics show the versatility of the proposal.
\citep{hensman2015} proposed a Dirichlet Process Gaussian Process (DP-GP) model for clustering structured time-series data, effectively capturing both intra-group and inter-group variability. However, their approach focuses on clustering from a static perspective, without explicitly addressing the temporal dynamics underlying the time-series data.
Importantly, an extension to nonlinear (ODEs) and a significant improvement in computational complexity for latent force modelling have been achieved in~\citep{hartikainen2011, hartikainen2012}. However, their application to dynamic clustering with nonlinear mappings to observations given by GP priors faces identifiability and scalability issues due to the nested dependencies over complex GP structures.

\section{Conclusion}
In this paper, we have presented a Bayesian nonparametric approach to learning switching linear dynamical models for describing the evolution of an unbounded number of time series clusters. We used Gaussian process priors to model the variability both in amplitude and temporal alignment of the elements within each cluster. We have demonstrated the utility and adaptability of our approach in different real scenarios and its ability to properly unveil some of the mechanisms generating the data, thereby enabling their estimation. Hyperparameters of the model are easily interpretable, providing actionable criteria to guide the learning process and to qualify the meaning of the results.

Offline and online variational inference methods, with good scalability properties, have been devised\footnote{An open-source implementation is available online (code will be made available upon acceptance for anonymity)} 
and applied to ECG heartbeat clustering, respiration estimation, directional wave-spectra clustering, and pedestrian trajectory clustering. In the case of ECG clustering, HDP-GPC shows better results than previous state-of-the-art proposals by evolving each cluster over time, thus avoiding the proliferation of new clusters in static approaches to preserve good performance. The online method has been proven suitable for processing data streams, and the number of time-series segments to be processed at a time can be increased according to resource availability. Beyond ECG, the application to directional wave-spectra analysis shows that the model can exploit multi-output observations when several directional channels describe a common evolving physical regime providing compact and interpretable sea-states. In pedestrian trajectory clustering, an static version of the model proves versatile when the sample order is not meaningful, recovering interpretable routes. Still, the warping computation can be particularly costly and arises as the main barrier to real-time processing, demanding a more efficient approach like the path-wise sampling of the warp functions proposed in~\citep{mikheeva2022}. A possible direction of future research is a fully coupled multi-output formulation in which cross-output correlations are represented explicitly through block-structured LDS parameters or multi-output GP kernels, rather than through the conditionally independent weighted likelihood used in this paper. Another interesting line of work would be to generalise the Markovian switching by including additional dependencies on observations or continuous latent states, endowing the model with more versatility and interpretability~\citep{linderman2017}.

Ultimately, the present proposal represents a flexible, Bayesian nonparametric approach for describing complex dynamical phenomena through some temporal substructures that evolve according to linear dynamics.


\bibliographystyle{ACM-Reference-Format}
\bibliography{biblio}

\clearpage

\appendix

\section{Derivation of Variational Updates}
\label{ap:variational_updates}
In this appendix, we complement the variational updates provided in section~\ref{sec:variational_inference}.
In equation~\eqref{eq:q_S} the terms are defined for $n>0$, 
\begin{align}
\log \zeta_{njk} &= \mathbb{E}[\log \pi_{jk}] - \frac{1}{2} \mathbb{E}[\log |K^{x_n^k}_{\mathbf{g}^k_n,\mathbf{g}^k_n}|] - \frac{|\mathbf{y}_n|}{2}\log (2\pi) - \frac{1}{2} \mathbb{E}_{{x_n^k},\mathbf{g}^k_n} [(\mathbf{y}_n-m_{x_n^k}(\mathbf{g}^k_n))^T  K^{x_n^k-1}_{\mathbf{g}^k_n,\mathbf{g}^k_n}(\mathbf{y}_n-m_{x_n^k}(\mathbf{g}^k_n))] \label{eq:zeta_term}
\end{align}
where, 
\begin{equation}
\begin{array}{ll}
     m_{x_n^k}(\mathbf{g}^k_n) & = K^{\theta ^{(k)}}_{\mathbf{g}^k_n,\mathbf{t}_n}K^{\theta ^{(k)}-1}_{\mathbf{t}_n,\mathbf{t}_n}\mathbf{x}^{(k)}_n\\
     K^{x_n^k}_{\mathbf{g}^k_n,\mathbf{g}^k_n} &  = K^{\theta ^{(k)}}_{\mathbf{g}^k_n,\mathbf{g}^k_n}-K^{\theta ^{(k)}}_{\mathbf{g}^k_n,\mathbf{t}_n}K^{\theta ^{(k)}-1}_{\mathbf{t}_n,\mathbf{t}_n}K^{\theta ^{(k)}}_{\mathbf{t}_n,\mathbf{g}^k_n}.
\end{array}
\end{equation}
Where the expectation over $\mathbf{g}_n^k$ results from a point estimate approximation of the warping function (see below).

In equation~\ref{eq:q_pi} the expectations $\mathbb{E}_v[\beta_k]$ are computed from equation (\ref{eq:stick_breaking}). The parameterization of $q^*(\pi)$ allows the computation of the expectation involved in equation (\ref{eq:zeta}) as
\begin{equation}
    \mathbb{E}\left[\log \pi_{j k}\right] = \psi(\kappa_{j k}) - \psi\left(\sum_{k=1}^{K+1}\kappa_{j k}\right),
\label{eq:transition_parameterized}
\end{equation}
where $\psi(\cdot)$ is the digamma function.

For the \emph{maximum a posteriori} (MAP) approach of the warp functions, we maximise
\begin{align}
\log q^*(\mathbf{g}_n^k|\mathbf{y}_n,\mathbf{t}_n) &= \log p(\mathbf{y}_n|m_{x_n^k}(\mathbf{g}_n^k), K^{x_n^k}_{\mathbf{g}^k_n,\mathbf{g}^k_n}) + \log p(\mathbf{g}^k_n|\mathbf{t}_n) + \mathrm{const.}
\label{eq:maximizing_g}
\end{align}
This time warping is a realization $\mathbf{g}^k_n\sim \mathcal{N}(m_{\vartheta}^k(\mathbf{t}_n),K^{\vartheta ^k}_{\mathbf{t}_n,\mathbf{t}_n})$ of the corresponding GP. The parameterisation of each $\mathbf{g}^k_n$ is ensured using a set of auxiliary variables $\mathbf{a}^k_n\in \mathbb{R}^Q$ such that $g^k_{n,q}=Q\sum^q_{i=1}[\operatorname{softmax}(\mathbf{a}^k_n)]_i$, with $\mathbf{a}^k_n\sim \mathcal{N}(\mathbf{0}, I)$, and appropriate hyperparameters $\vartheta$ to ensure smoothness.

For the update of the dynamical parameters, $\mathcal{MN}(M^k_A,W^k_A,V^k_A)$ denotes a matrix-normal prior with mean matrix $M^k_A$, left covariance $(V^k_A)^{-1}$ and right covariance $W^k_A$, and $\mathcal{IW}(\nu_0,S^k_\omega)$ denotes an inverse-Wishart prior with $\nu_0$ degrees of freedom and scale matrix $S^k_\omega$. The log of the optimised factor is given by
\begin{align}
\log q^*(A^k,\Sigma^k_\omega) &= \sum_{m=1}^{n_k}\mathbb{E}_{\mathbf{f}}[\log p(\mathbf{f}^k_{m}|\mathbf{f}^k_{m-1},A^k,\Sigma^k_\omega)] + \log p(A^k,\Sigma^k_\omega) + \mathrm{const.}
\end{align}
where the index $m$ runs through the elements of the cluster $k$. The optimised factor can be derived as a posterior distribution \cite{fox_2011}
\begin{align}
q^*(A^k,\Sigma^k_{\omega}) &=\mathcal{MN}(\Psi_{m,m-1}\Psi^{-1}_{m-1,m-1},\Sigma^k_{\omega},\Psi_{m-1,m-1}) \mathcal{IW}(N_k+\nu_0,\Psi_{m|m-1}+S^k_\omega)
\end{align}
where
\begin{align}
\Psi_{m-1,m-1} &= \mathbb{E}[\mathbf{f}^k_{m-1}\mathbf{f}^{kT}_{m-1}]+V^k_A, \nonumber \\
\Psi_{m,m-1} &= \mathbb{E}[\mathbf{f}^k_{m}\mathbf{f}_{m-1}^{kT}]+M^k_AV^k_A, \nonumber \\
\Psi_{m,m} &= \mathbb{E}[\mathbf{f}^k_{m}\mathbf{f}_{m}^{kT}]+M^k_AV^k_AM_A^{kT}, \nonumber \\
\Psi_{m|m-1} &=\Psi_{m,m}-\Psi_{m,m-1}\Psi_{m-1,m-1}^{-1}\Psi_{m,m-1}^T \label{eq:dynamical_param_estimators}
\end{align}
We denote the new posteriors by $\mathcal{MN}(\widetilde{M}^k_A,\widetilde{W}^k_A,\widetilde{V}^k_A)$ and $\mathcal{IW}(\tilde{\nu}_{0},\widetilde{S}^k_\omega)$ where $\tilde{\nu}_0=N_k+\nu_0$. The optimized factor for the parameters of (\ref{eq:emission_trans}) can be obtained analogously, resulting in $\mathcal{MN}(\widetilde{M}^k_C,\widetilde{W}^k_C,\widetilde{V}^k_C)$ and $\mathcal{IW}(\tilde{\varsigma}_{0},\widetilde{S}^k_\epsilon)$ where $\tilde{\varsigma}_0=N_k+\varsigma_0$.

For the Kalman filtering and smoothing equations, we distinguish it as a two-step process. As a result of the forward equations, we obtain
\begin{equation}
q(\mathbf{f}^k_n)=\mathcal{N}(\mathbf{f}^k_n|\mu_{\mathbf{f}^k_n},\Sigma_{\mathbf{f}^k_n})
\end{equation}
where
\begin{align}
\mu_{\mathbf{f}^k_n} &\triangleq \mathbb{E}[\mathbf{f}_n^k]=\widetilde{M}_A^k \mu_{\mathbf{f}^k_{n-1}} +  H_n^k (\mathbf{y}_{n} - \hat{K}_n^k\widetilde{M}_C^k\widetilde{M}_A^k\mu_{\mathbf{f}^k_{n-1}}) \nonumber \\
\Sigma_{\mathbf{f}^k_n} &\triangleq \cov(\mathbf{f}_n^k) = (I-H_n^k\hat{K}_n^k\widetilde{M}_C^k) P_{n-1}^k, 
\label{eq:forward_message}
\end{align}
with
\begin{align}
H_n^k &=P_{n-1}^k (\hat{K}_n^k \widetilde{M}_C^k)^T (K^{\theta^k}_{\mathbf{g}_n^k,\mathbf{g}_n^k}-\hat{K}_n^k K^{\theta^k}_{\mathbf{t}_n,\mathbf{g}_n^k}+ \hat{K}_n^k \Phi_n^k \hat{K}_n^k)^{-1}, \nonumber \\
\Phi_n^k &= (\widetilde{M}_C^kP_{n-1}^k\widetilde{M}_C^{kT}+\widetilde{S}_{\epsilon}^k)\oslash r_{nk}, \nonumber \\
P_{n-1}^k &= \widetilde{M}_A^k \Sigma_{\mathbf{f}_{n-1}^k} \widetilde{M}_A^{kT} + \widetilde{S}_{\omega}^k, \nonumber \\
\hat{K}_n^k &= K^{\theta^k}_{\mathbf{g}_n^k,\mathbf{t}_n} K^{\theta^k -1}_{\mathbf{t}_n,\mathbf{t}_n}.
\end{align}
In the above equations, the index $n$ represents $n_k$ when it comes with some descriptor of cluster $k$. The initial condition for this message scheme is $q^*(\mathbf{f}_0^k)=\mathcal{N}(m_{\theta^k}(t_0),K^{\theta^k}_{\mathbf{t}_0,\mathbf{t}_0})$. The matrix $H_n^k$ is the well-known Kalman gain matrix. The symbol $\oslash$ denotes the element-wise division. The matrix $\hat{K}_n^k$ represents the KL-optimal projection introduced in equation (\ref{eq:seeger_approach}), to compute the posteriors in locations other than those of the observations, allowing us to efficiently reduce the computation of latent GPs to a small set of inducing points \cite{rasmussen_gaussian_2006}. Note that in the limit $r_{nk}\rightarrow 0$, the Kalman gain goes to zero, as it is expected.

Now running the smoothing step to compute the smoothed posterior
\begin{equation}
q^*(\mathbf{f}_n^k)=\mathcal{N}(\mathbf{f}_n^k|\tilde{\mu}_{\mathbf{f}_n^k},\widetilde{\Sigma}_{\mathbf{f}_n^k})
\label{eq:forward_backward}
\end{equation}
where
\begin{align}
\tilde{\mu}_{\mathbf{f}_n^k} &\triangleq \mu_{\mathbf{f}^k_n} + J_n^k (\tilde{\mu}_{\mathbf{f}^k_{n+1}} - \widetilde{M}_A^k \mu_{\mathbf{f}^k_n}) \nonumber \\
\widetilde{\Sigma}_{\mathbf{f}_n^k} &\triangleq \Sigma_{\mathbf{f}^k_n} + J_n^k (\widetilde{\Sigma}_{\mathbf{f}^k_{n+1}} - P_n^k) J_n^{kT}
\end{align}
with
\begin{equation}
J_n^k= \Sigma_{\mathbf{f}^k_n}\widetilde{M}_A^{kT} (P_n^{k})^{-1}.
\end{equation}
The $q^*(\mathcal{X}|\mathcal{F})$ factor can be obtained by projection under the respective dynamic equations of each cluster
\begin{equation}
q^*(\mathbf{x}_n^k|\tilde{\mathbf{f}}_n^k) = \mathcal{N}(\mathbf{x}_n^k|\widetilde{M}_C^k {\tilde{\mathbf{f}}_n^k},\widetilde{S}_{\epsilon}^k),
\end{equation}
and $q^*(\mathcal{X})$ can be computed by marginalization
\begin{equation}
q^*(\mathbf{x}_n^k) = \mathcal{N}(\mathbf{x}_n^k|\tilde{\mu}_{\mathbf{x}_n^k},\widetilde{\Sigma}_{\mathbf{x}_n^k}),
\end{equation}
where
\begin{align}
\tilde{\mu}_{\mathbf{x}_n^k} &\triangleq \widetilde{M}_C^k \tilde{\mu}_{\mathbf{f}^k_n} \nonumber \\
\widetilde{\Sigma}_{\mathbf{x}_n^k} &\triangleq \widetilde{S}_{\epsilon}^{k} + \widetilde{M}_C^{kT}\widetilde{\Sigma}_{\mathbf{f}^k_n}\widetilde{M}_C^k
\end{align}
\section{Derivation of the Variational Objective}
\label{ap:ELBO}

The variational objective can be decomposed as
\begin{equation}
\mathcal{L} =\mathcal{L}_{OBS} + \mathcal{L}_{HDP} + \mathcal{L}_{W}+ \mathbb{H}[q(\mathcal{S}\,)],
\label{eq:lower_bound_ap}
\end{equation}
which, due to the factorisation of the variational family, is given by
\begin{align}
    \mathcal{L}_{OBS} &= \mathbb{E}_{q}\!\left[\log p\!\left(\mathcal{Y}| \mathcal{X},\mathcal{G}, \mathcal{S}\right) + \log \frac{p(\mathcal{\phi})}{q(\mathcal{\phi})} + \log \frac{p(\mathcal{X}|\mathcal{F},\phi)p(\mathcal{F}|\phi)}{q(\mathcal{X} |\mathcal{F})q(\mathcal{F})} \right], \\ 
    \mathcal{L}_{HDP} &= \mathbb{E}_{q}\! \left[\log\frac{p(\mathcal{S}|\pi)p(\pi|v,\alpha)}{ q\!\left(\pi\right)} + \log \frac{p(v|\gamma)}{q(v|\lambda,\eta)} \right], \label{eq:L_HDP}
\end{align}
where we have omitted the $*$ superscript on the $q$ distributions. $\mathbb{H}[q(\mathcal{S})]$ is defined as the Shannon entropy of the distribution $q(\mathcal{S})$. Note that $\mathcal{G}$ is inferred via MAP and treated as a point estimate, so the entropy is excluded from testing for convergence, as $\delta_{\mathcal{G}^*}(\mathcal{G})$ is always degenerate. By including a MAP estimate for $\mathcal{G}$, the global approximation is locally optimised, with the expectation of $p(\mathcal{M}\, |\, \mathcal{G}\,)$ remaining fully variational.

The inclusion of $q(v)$ as a variational factor results in a source of intractability at computing $\mathcal{L}_{HDP}$ \cite{hughes_2015}. Specifically,
\begin{align}
\mathbb{E}_{q}\!\left[\log\frac{p(\mathcal{S}|\pi)p(\pi|v,\alpha)}{q\!\left(\pi\right)}\right] &= (K+1)\mathbb{E}[c_D(\alpha\beta)] + \sum_{j=1}^{K+1} c_D(\kappa_j) + \sum_{k=1}^{K+1}\sum_{j=1}^{K+1} (N_{jk}+\alpha\mathbb{E}[\beta_k]-\kappa_{jk})\mathbb{E}[\log \pi_{jk}] \label{eq:trans_bound}
\end{align}
where expectations are taken with respect to all of the factors involved. The term $\mathbb{E}_q[c_D(\alpha\beta)]$ represents the expectation of the cumulant function of the Dirichlet and it has no closed form. A lower bound $c_D(\alpha \beta)\geq K \log\alpha + \sum_{k=1}^{K+1} \log v_k + \sum_{k=1}^{K+1} (K+1-k)\log(1-v_k)$ is proven to be valid for all $\alpha >0$ \cite{hughes_2015}, and it can be used as a surrogate bound in equation (\ref{eq:lower_bound_ap}). Expectations can now be tractably computed since they involve logs of Beta distributions. On the other hand, there is no closed-form update of the parameters $\lambda_{k}, \eta_{k}$, and they are updated through numerical optimization of $\lambda_{k}, \eta_{k} = \mathrm{argmax}_{\lambda_{k},\eta_{k}} \mathcal{L}_{HDP}$ \cite{hughes_2015}. The second term of $\mathcal{L}_{HDP}$ is easily computable as it only involves expectations over Beta distributions:
\begin{align}
\mathbb{E}_{q}\!\left[\log \frac{p(v | \gamma)}{q(v | \lambda, \eta)} \right] &= \sum_{k=1}^{K+1}c_{B}(1,\gamma) - c_B(\lambda_{k}\eta_{k}, (1-\lambda_{k})\eta_{k}) + (1-\lambda_{k}\eta_{k})\mathbb{E}_q[\log v_{k}] + (\gamma - (1-\lambda_{k})\eta_{k})\mathbb{E}_q[\log (1-v_{k})]\label{eq:stick_bound}
\end{align}
where $c_{B}(\cdot)$ is the cumulant function of the Beta distribution i.e. a simplified two argument case of $c_D(\cdot)$. The expectations of the Beta distributions are $\mathbb{E}_q[\log v_{k}] = \psi(\lambda_{k}\eta{k}) - \psi(\eta_{k})$ and $\mathbb{E}_q[\log (1-v_{k})] = \psi((1-\lambda_{k})\eta{k}) - \psi(\eta_{k})$ where $\psi(\cdot)$ is the digamma function. Collecting each terms related to data generation to evaluate $\mathcal{L}_{OBS}$, we have:
\begin{align}
\mathbb{E}_{q}\!\left[\log p(\mathcal{Y} | \mathcal{X}, \mathcal{G}, \mathcal{S}) \right] &= \sum_{k=1}^{K+1}\sum_{n=1}^{N} r_{nk}\log \zeta_{nk} 
\end{align}
where,
\begin{align}
\log \zeta_{nk} &= - \frac{1}{2} \log |K^{x_n^k}_{\mathbf{g}^k_n,\mathbf{g}^k_n}| - \frac{|\mathbf{y}_n|}{2}\log (2\pi) - \frac{1}{2} \mathbf{y}_n^{T} K^{{x_n^k}-1}_{\mathbf{g}^k_n,\mathbf{g}^k_n} \mathbf{y}_n + \mathbf{y}_n^{T} K^{{x_n^k}-1}_{\mathbf{g}^k_n,\mathbf{g}^k_n} \hat{K}_{\mathbf{g}^k_n}^{k}\mathbb{E}[\mathbf{x}^{k}_n] \nonumber \\
& - \frac{1}{2} tr\left[\hat{K}_{\mathbf{g}^k_n}^{kT}K^{{x_n^k}-1}_{\mathbf{g}^k_n,\mathbf{g}^k_n}\hat{K}_{\mathbf{g}^k_n}^{k}\left(\mathbb{E}[\mathbf{x}^{k}_n]\mathbb{E}[\mathbf{x}^{k}_n]^{T}+\text{cov}(\mathbf{x}^{k}_n)\right)\right]  \label{eq:example_bound}
\end{align}
where, 
\begin{equation}
\begin{array}{rl}
     \hat{K}_{\mathbf{g}^k_n}^{k} &= K^{\theta^{(k)}}_{\mathbf{g}_n^k,\mathbf{t}_n} K^{\theta^k -1}_{\mathbf{t}_n,\mathbf{t}_n} \\
     m_{x_n^k}(\mathbf{g}^k_n) & = \hat{K}_{\mathbf{g}^k_n}^{k}\mathbf{x}^{k}_n \\
     K^{x_n^k}_{\mathbf{g}^k_n,\mathbf{g}^k_n} &  = K^{\theta ^{(k)}}_{\mathbf{g}^k_n,\mathbf{g}^k_n}-\hat{K}_{\mathbf{g}^k_n}^{k}K^{\theta ^{(k)}}_{\mathbf{t}_n,\mathbf{g}^k_n}
\end{array}
\end{equation}
Then, the auto-regressive factor for the latent process is
\begin{align}
\mathbb{E}_{q}\!\left[\log \frac{p(\mathcal{F} | \phi)}{q(\mathcal{F})} \right] & = \frac{1}{2}\sum_{k=1}^{K} N_{k} \log \frac{|S_{\omega}^{k-1}|}{|\tilde{\Sigma}_{\mathbf{f}_{m}^{k}}^{-1}|} + \sum_{k=1}^{K} \sum_{m=1}^{N_{k}} tr(\tilde{\Sigma}_{\mathbf{f}_{m}^{k}}^{-1}\tilde{\mu}_{\mathbf{f}_{m}^{k}}\tilde{\mu}_{\mathbf{f}_{m}^{k}}^{T}) - tr(S_{\omega}^{k-1}\mathbb{E}[\mathbf{f}_{m}^{k}\mathbf{f}_{m}^{k T}]) \nonumber \\
& + 2tr(S_{\omega}^{k-1}M_{A}^{k}\mathbb{E}[\mathbf{f}_{m-1}^{k}\mathbf{f}_{m}^{k T}])- tr(M_{A}^{k T}S_{\omega}^{k-1}M_{A}^{k}\mathbb{E}[\mathbf{f}_{m-1}^{k}\mathbf{f}_{m-1}^{k T}])
\end{align}
where the expectations are evaluated using the estimated covariances of the latent variables. The factor associated with the emission process $\mathcal{X}$ is analogous. The warp regularisation factor descomposed as 
\begin{equation}
    \mathcal{L}_{\mathcal{G}}(\hat{\mathcal{G}}\,) = \sum_{k=1}^{K+1}\sum_{m=1}^{N_{k}} \log p(\mathbf{g}_{m}^{k}\mid \mathbf{t}_{n})
\end{equation}
is computed as
\begin{equation}
\mathcal{L}_{\mathcal{G}}(\hat{\mathcal{G}}\,) =\sum_{k=1}^{K} \sum_{m=1}^{N_{k}} -\frac{|\mathbf{t}_{n}|}{2}\log(2\pi) + \frac{1}{2}\log|K^{\boldsymbol{\vartheta} ^k -1}_{\mathbf{t}_n,\mathbf{t}_n}|  - \frac{1}{2} \mathbf{g}_n^{k\top} K^{\boldsymbol{\vartheta} ^k -1}_{\mathbf{t}_n,\mathbf{t}_n} \mathbf{g}_n^{k} + \mathbf{g}_n^{k\top} K^{\boldsymbol{\vartheta} ^k -1}_{\mathbf{t}_n,\mathbf{t}_n}  m_{\boldsymbol{\vartheta}}^k(\mathbf{t}_n) - \frac{1}{2} m_{\boldsymbol{\vartheta}}^k(\mathbf{t}_n) K^{\boldsymbol{\vartheta} ^k -1}_{\mathbf{t}_n,\mathbf{t}_n} m_{\boldsymbol{\vartheta}}^k(\mathbf{t}_n)
\end{equation}
Finally, the factor involving the dynamical parameters is computed as
\begin{align}
\mathbb{E}_{q}\!\left[\log \frac{p(\phi)}{q(\phi)} \right] &= \sum_{k=1}^{K}\mathbb{E}_{q}\!\left[\log \frac{p(A^{k},\Sigma_{\omega}^{k})p(C^{k},\Sigma_{\epsilon}^{k})}{q(A^{k},\Sigma_{\omega}^{k})q(C^{k},\Sigma_{\epsilon}^{k})} \right]
\end{align}
where, 
\begin{align}
\mathbb{E}_{q}\!\left[\log p(A^{k},\Sigma_{\omega}) \right] &= c_{MN}(M_{A}^{k},W_{A}^{k},V_{A}^{k}) + c_{IW}(\nu_{0}^{k},S_{\omega}^{k}) + \frac{\nu_{0}^{k}-1}{2}\log |\Sigma_{\omega}^{k-1}| - \frac{1}{2}tr(\mathbb{E}[A^{kT}\Sigma_{\omega}^{k-1}A^{k}]V_{A}^{k}) \nonumber \\
&+ tr(\mathbb{E}[\Sigma_{\omega}^{k-1}A^{k}]V_{A}^{k}M_{A}^{k})  - \frac{1}{2}tr(\mathbb{E}[\Sigma_{\omega}^{k-1}](S_{\omega}^{k-1} + V_{A}^{kT}M_{A}^{k}V_{A}^{k}))
\end{align}
with, 
\begin{align}
\mathbb{E}[\Sigma_{\omega}^{k-1}] &= \frac{1}{\nu-D+1}S_{\omega}^{k-1} \\
\mathbb{E}[A^{k}] &= M_{A}^{k}\\
\operatorname{cov}(A^{k}) &= \frac{1}{\nu-D+1} (V_{A}^{k-1} \times S_{\omega}^{k-1}) \\
c_{MN}(M_{A}^{k},W_{A}^{k},V_{A}^{k}) &= -\frac{D^2}{2}\log 2\pi + \frac{D}{2}\log|V_{A}^{k-1}| \\
c_{IW}(\nu_{0}^{k},S_{\omega}^{k}) &= -\frac{D(D-1)}{4}\log\pi - \frac{D}{2}(\log 2)\nu_{0}^{k} - \sum_{d=1}^{D} \log \Gamma\left(\frac{\nu_{0}^{k}+1-d}{2}\right) + \frac{\nu_{0}^{k}}{2}\log|S_{\omega}^{k-1}|
\end{align}
To compute the difference with the variational factor $\mathbb{E}[\log q(A^{k},\Sigma_{\omega}^{k})]$ the same evaluation is used but replacing the variational distribution $\mathcal{MN}(\widetilde{M}^k_A,\widetilde{W}^k_A,\widetilde{V}^k_A)\mathcal{IW}(\tilde{\nu}_{0},\widetilde{S}^k_\omega)$ with the estimated form of the mean and covariance. The factors associated with $\mathbb{E}[\log p(C^{k},\Sigma_{\epsilon}^{k})]$ and $\mathbb{E}[\log q(C^{k},\Sigma_{\epsilon}^{k})]$ are analogous.

\section{Additional MIT-BIH Clustering and Runtime Results}
\label{app:mitbih_results}

This appendix reports the complete ECG clustering results on the MIT-BIH Arrhythmia Database. The main text focuses on cluster-count reduction and global purity, while the present appendix includes additional external clustering metrics, classical and deep clustering baselines, per-record results for the multi-output offline version of HDP-GPC, the AAMI-level confusion matrix, and the empirical runtime analysis.

\subsection{Comparison with classic and deep clustering baselines}

Table \ref{tab:mitbih_global_appendix} summarises the global comparison with previous heartbeat clustering methods and with the additional unsupervised baselines. For the DTW, deep clustering and HDP-GPC methods, values are reported as mean $\pm$ standard deviation across the 48 records. The number of clusters corresponds to the total number of clusters generated across all records. For HDP-GPC, the table reports record-wise mean metrics, while the global beat-weighted purity of the multi-output offline method is 97.53\%.

\begin{table}
\centering{
\scriptsize
\setlength{\tabcolsep}{3pt}
\renewcommand{\arraystretch}{1.1}
\caption{Global clustering results on the MIT-BIH Arrhythmia Database.}
\label{tab:mitbih_global_appendix}
\resizebox{\textwidth}{!}{
\begin{tabular}{l c r c c c c c c c}
\hline\hline
Method  & Clusters & Acc./Purity & ARI & AMI & NMI & Homo. & Comp. & V-meas. \\ \hline
Lagerholm et al. \cite{lagerholm2000}  & 1200 & 0.985 & -- & -- & -- & -- & -- & -- \\
Castro et al. \cite{castro2015}  & 629 & 0.972 & -- & -- & -- & -- & -- & -- \\
DTW+k-medoids \cite{Sakoe_1978, Kaufman_1990}  & 119 & $0.943\pm0.063$ & $0.224\pm0.340$ & $0.221\pm0.306$ & $0.222\pm0.305$ & $0.288\pm0.322$ & $0.232\pm0.330$ & $0.222\pm0.305$ \\
DTW+hierarchical-average \cite{Sakoe_1978, Murtagh_2012} & 107 & $0.934\pm0.087$ & $0.367\pm0.399$ & $0.340\pm0.372$ & $0.341\pm0.372$ & $0.355\pm0.373$ & $0.506\pm0.413$ & $0.341\pm0.372$ \\
CAE+DBSCAN \cite{Masci_2011, Ester_1996}  & 168 & $0.946\pm0.064$ & $0.373\pm0.308$ & $0.323\pm0.270$ & $0.325\pm0.269$ & $0.490\pm0.302$ & $0.287\pm0.243$ & $0.325\pm0.269$ \\
CAE+DPGMM \cite{Masci_2011, Antoniak_1974}  & 906 & $0.979\pm0.041$ & $0.271\pm0.326$ & $0.298\pm0.281$ & $0.301\pm0.281$ & $0.754\pm0.278$ & $0.225\pm0.237$ & $0.301\pm0.281$ \\
DEC \cite{Xie_2016} & 128 & $0.958\pm0.059$ & $0.375\pm0.421$ & $0.352\pm0.383$ & $0.353\pm0.382$ & $0.461\pm0.371$ & $0.364\pm0.409$ & $0.353\pm0.382$ \\
HDP-GPC MO offline  & 342 & $0.976\pm0.045$ & $0.434\pm0.356$ & $0.435\pm0.312$ & $0.437\pm0.312$ & $0.665\pm0.350$ & $0.414\pm0.319$ & $0.437\pm0.312$ \\ \hline\hline
\end{tabular}}}
\vspace{0.1cm}

\begin{minipage}{0.98\textwidth}
\footnotesize
Values reported as $\mu\pm\sigma$ correspond to the arithmetic mean and
standard deviation computed across the 48 record-level scores of the
MIT-BIH Arrhythmia Database. Thus, each record contributes one value to
Acc./Purity, ARI, AMI, NMI, homogeneity, completeness and V-measure,
independently of the number of beats in the record. Cluster counts report
the total number of clusters generated across all records. The values from
\cite{lagerholm2000} and \cite{castro2015} are reproduced as reported in
the corresponding references; no record-wise dispersion was available for
those methods.
\end{minipage}
\end{table}

The deep clustering baselines provide competitive label-matching accuracy, particularly CAE+DPGMM~\cite{Masci_2011, Antoniak_1974}. However, HDP-GPC obtains the strongest record-wise ARI, AMI, NMI, completeness, and V-measure among the compared unsupervised methods, while also producing interpretable GP morphology representatives and inferring the number of clusters without fixing it a priori. For the additional baselines, all methods were applied independently to each record using the same QRS segmentation and feature support as HDP-GPC. The number of clusters for DTW+k-medoids~\cite{Sakoe_1978, Kaufman_1990} and DTW+hierarchical clustering~\cite{Sakoe_1978, Murtagh_2012} was selected by maximizing the silhouette score~\cite{Rousseeuw_1987} over candidate values of $K$, using the precomputed DTW distance matrix and average-linkage agglomerative clustering as the selection model. The search was performed over $K\in [2,K_{max}]$ with $K_{max}=20$ by default. DBSCAN~\cite{Ester_1996} parameters were selected by a grid search over \texttt{min\_samples} values $\{3,5,10\}$ and candidate $\epsilon$ values obtained from quantiles of the k-nearest-neighbour distance curve; the selected configuration was the one maximizing the silhouette score over the non-noise samples. The CAE~\cite{Masci_2011} architecture was a one-dimensional convolutional autoencoder with three encoder blocks using 16, 32, and 64 filters, followed by a fully connected latent representation of dimension 5. The decoder used symmetric transposed-convolution layers. The CAE was trained for 300 epochs using Adam~\cite{kingma2015adam} with learning rate $10^{-4}$, weight decay $10^{-5}$, and batch size 128. DEC~\cite{Xie_2016} was initialized from the trained CAE encoder; cluster centres were initialized using k-means in the latent space, with $K$ selected automatically by silhouette maximization up to $K_{max}=20$. The DEC objective was optimized for 300 epochs using Adam with learning rate $10^{-5}$ minimizing the KL divergence between the soft assignments and the sharpened target distribution.

\subsection{Per-record results for the multi-output offline method}

Table \ref{tab:mitbih_metrics_mo_appendix} reports the complete per-record results for the multi-output offline version of HDP-GPC, denoted by $N_{MO}$. This is the configuration used as the main multi-output result in the paper. The method processes both ECG leads jointly by means of the SNR-weighted likelihood. For clarity, the table includes two complementary summary rows. The row \emph{Total/mean} reports the descriptive mean of the per-record metrics together with the total number of beats and inferred clusters, and is therefore representative of the typical record-level behaviour of the model. The row \emph{Global}, by contrast, reports the corresponding metrics computed from the database-wide contingency table obtained by pooling all beats across all records before evaluation. This second summary preserves the full class imbalance of MIT-BIH and is therefore the most faithful characterisation of clustering behaviour at the database scale, but it is also more conservative for chance-corrected and entropy-based metrics.

The difference between the two rows is informative. Purity is nearly unchanged in the two summaries (0.976 in \emph{Total/mean} versus 0.975 in \emph{Global}), showing that the method remains highly accurate both at the record level and across the full database. However, metrics such as NMI and ARI are noticeably lower in the \emph{Global} row (0.392 and 0.028) than in the \emph{Total/mean} row (0.437 and 0.434). This reflects the strong label imbalance of MIT-BIH, where normal beats dominate the database, and the fact that the global contingency table is heavily influenced by the mismatch between coarse physiological labels and the finer morphology-specific families recovered by the model. For this reason, per-record values are often more informative when interpreting clustering quality in balanced recordings: for example, records 106, 119, 215, and 221 obtain ARI values of 0.891, 1.000, 0.973, and 0.962, respectively. Likewise, the combination of higher homogeneity than completeness indicates that the model tends to produce clusters that are nearly pure, while still splitting coarse annotation labels into several morphology-specific subfamilies. This behaviour is expected for morphology clustering, because MIT-BIH labels encode activation origin and conduction route rather than a purely morphological taxonomy.

\begin{table}
\centering{
\scriptsize
\setlength{\tabcolsep}{3.2pt}
\renewcommand{\arraystretch}{1.05}
\caption{Per-record clustering metrics for the multi-output offline HDP-GPC method.}
\label{tab:mitbih_metrics_mo_appendix}
\begin{tabular}{c r r c c c c c c}
\hline\hline
Record & Beats & Clusters & Purity & NMI & ARI & Homo. & Comp. & M-F1. \\ \hline
100 & 2272 & 2 & 0.985 & 0.092 & 0.056 & 0.048 & 1.000 & 0.664 \\
101 & 1865 & 2 & 0.998 & 0.584 & 0.571 & 0.413 & 1.000 & 0.666 \\
102 & 2187 & 10 & 0.991 & 0.452 & 0.328 & 0.872 & 0.305 & 0.932 \\
103 & 2084 & 4 & 0.999 & 0.000 & -0.002 & 0.002 & 0.000 & 0.500 \\
104 & 2229 & 15 & 0.918 & 0.496 & 0.335 & 0.785 & 0.363 & 0.497 \\
105 & 2572 & 4 & 0.986 & 0.496 & 0.661 & 0.619 & 0.413 & 0.566 \\
106 & 2027 & 4 & 0.992 & 0.768 & 0.891 & 0.919 & 0.660 & 0.989 \\
107 & 2137 & 6 & 0.997 & 0.236 & 0.095 & 0.850 & 0.137 & 0.972 \\
108 & 1763 & 2 & 0.993 & 0.616 & 0.759 & 0.581 & 0.656 & 0.359 \\
109 & 2532 & 8 & 0.998 & 0.224 & 0.155 & 0.903 & 0.128 & 0.653 \\
111 & 2124 & 2 & 1.000 & 1.000 & 1.000 & 1.000 & 1.000 & 1.000 \\
112 & 2539 & 3 & 0.999 & 0.000 & -0.001 & 0.009 & 0.000 & 0.500 \\
113 & 1794 & 3 & 0.997 & 0.016 & 0.004 & 0.254 & 0.008 & 0.721 \\
114 & 1879 & 4 & 0.991 & 0.663 & 0.725 & 0.694 & 0.635 & 0.388 \\
115 & 1952 & 3 & 1.000 & 0.000 & 0.000 & 1.000 & 0.000 & 1.000 \\
116 & 2411 & 7 & 0.999 & 0.356 & 0.181 & 0.965 & 0.218 & 0.665 \\
117 & 1535 & 4 & 0.999 & 0.001 & -0.000 & 0.109 & 0.001 & 0.500 \\
118 & 2278 & 5 & 0.958 & 0.311 & 0.247 & 0.196 & 0.747 & 0.659 \\
119 & 1987 & 3 & 1.000 & 0.997 & 1.000 & 1.000 & 0.993 & 1.000 \\
121 & 1863 & 4 & 0.999 & 0.091 & 0.057 & 0.501 & 0.050 & 0.667 \\
122 & 2476 & 1 & 1.000 & 1.000 & 1.000 & 1.000 & 1.000 & 1.000 \\
123 & 1518 & 3 & 1.000 & 0.089 & 0.041 & 1.000 & 0.047 & 1.000 \\
124 & 1619 & 9 & 0.975 & 0.306 & 0.187 & 0.542 & 0.213 & 0.409 \\
200 & 2601 & 16 & 0.978 & 0.674 & 0.814 & 0.836 & 0.565 & 0.492 \\
201 & 1963 & 10 & 0.974 & 0.551 & 0.351 & 0.776 & 0.428 & 0.533 \\
202 & 2136 & 7 & 0.982 & 0.246 & 0.134 & 0.510 & 0.162 & 0.781 \\
203 & 2980 & 7 & 0.970 & 0.413 & 0.366 & 0.717 & 0.290 & 0.378 \\
205 & 2656 & 10 & 0.995 & 0.716 & 0.824 & 0.908 & 0.592 & 0.664 \\
207 & 2332 & 9 & 0.897 & 0.642 & 0.740 & 0.760 & 0.556 & 0.686 \\
208 & 2955 & 17 & 0.969 & 0.711 & 0.799 & 0.908 & 0.585 & 0.703 \\
209 & 3005 & 6 & 0.891 & 0.167 & 0.206 & 0.108 & 0.377 & 0.734 \\
210 & 2649 & 9 & 0.981 & 0.561 & 0.670 & 0.769 & 0.441 & 0.384 \\
212 & 2747 & 3 & 1.000 & 0.900 & 0.908 & 1.000 & 0.818 & 1.000 \\
213 & 3250 & 15 & 0.961 & 0.498 & 0.472 & 0.745 & 0.375 & 0.680 \\
214 & 2262 & 12 & 0.996 & 0.481 & 0.355 & 0.932 & 0.325 & 0.496 \\
215 & 3363 & 16 & 0.999 & 0.786 & 0.973 & 0.968 & 0.661 & 0.665 \\
217 & 2208 & 11 & 0.984 & 0.709 & 0.603 & 0.933 & 0.571 & 0.960 \\
219 & 2154 & 12 & 0.996 & 0.403 & 0.334 & 0.854 & 0.263 & 0.623 \\
220 & 2048 & 3 & 0.954 & 0.007 & 0.045 & 0.012 & 0.005 & 0.488 \\
221 & 2427 & 7 & 1.000 & 0.835 & 0.962 & 1.000 & 0.716 & 1.000 \\
222 & 2483 & 4 & 0.830 & 0.005 & -0.017 & 0.004 & 0.007 & 0.227 \\
223 & 2605 & 11 & 0.952 & 0.510 & 0.592 & 0.676 & 0.409 & 0.323 \\
228 & 2053 & 14 & 0.999 & 0.425 & 0.196 & 0.978 & 0.271 & 0.666 \\
230 & 2256 & 4 & 1.000 & 0.007 & 0.001 & 1.000 & 0.004 & 1.000 \\
231 & 1571 & 4 & 0.999 & 0.904 & 0.926 & 0.982 & 0.838 & 0.749 \\
232 & 1780 & 2 & 0.776 & 0.002 & -0.002 & 0.001 & 0.034 & 0.291 \\
233 & 3078 & 21 & 0.994 & 0.618 & 0.786 & 0.961 & 0.455 & 0.534 \\
234 & 2753 & 4 & 0.987 & 0.404 & 0.519 & 0.319 & 0.551 & 0.829 \\ \hline
Total/mean & 109958 & 342 & 0.976 & 0.437 & 0.434 & 0.665 & 0.414 & 0.671 \\
Global     & 109958 & 342 & 0.975 & 0.392 & 0.028 & 0.928 & 0.249 & 0.636 \\ \hline\hline 
\end{tabular}
}
\vspace{0.2cm}

\centering
{\footnotesize Per-record clustering metrics for the multi-output offline method $N_{MO}$ using both leads of the MIT-BIH Arrhythmia Database. The row \emph{Total/mean} reports the descriptive mean of the per-record metrics together with the total number of beats and inferred clusters. The row \emph{Global} reports the corresponding metrics computed from the database-wide contingency table. NMI = Normalised Mutual Information, ARI = Adjusted Rand Index, Homo. = Homogeneity, Comp. = Completeness, M-F1 = Macro-F1.}
\end{table}

\subsection{AAMI-level confusion matrix}

Table \ref{tab:aami_confusion_appendix} reports the confusion matrix after collapsing the MIT-BIH labels into the five AAMI classes. Most errors occur between morphologies that are difficult to separate when atrial activity and rhythm information are intentionally excluded from the representation.

\begin{table}
\centering{
\scriptsize
\setlength{\tabcolsep}{6pt}
\renewcommand{\arraystretch}{1.1}
\caption{Global confusion matrix after AAMI label aggregation.}
\label{tab:aami_confusion_appendix}
\begin{tabular}{c r r r r r}
\hline\hline
 & N & S & V & F & Q \\ \hline
N & 90029 & 1101 & 152  & 67  & 199  \\
S & 406   & 1672 & 28   & 0   & 38   \\
V & 87    & 6    & 6934 & 74  & 46   \\
F & 68    & 1    & 46   & 662 & 0    \\
Q & 35    & 0    & 73   & 0   & 8232 \\ \hline\hline
\end{tabular}
}
\vspace{0.2cm}

\centering
{\footnotesize Columns contain the ground-truth labels and rows contain the dominant labels assigned to the clusters inferred by HDP-GPC.}
\end{table}

The confusion matrix reinforces the interpretation given in the main text. The proposed model is highly conservative with respect to within-cluster mixing, but its morphology-based representation may group together beats that differ mainly by atrial activity, rhythm context, or pacing artefacts. In particular, supraventricular and nodal beats are the main source of errors because the QRS complex alone does not always contain enough information to separate them from normal conduction. Paced and fusion beats constitute another relevant source of errors, especially in records 102, 104, and 217, where pacing spikes introduce sharp discontinuities that are not naturally represented by the squared-exponential kernel.

\subsection{Hyperparameters}
\label{app:hyperparams_mitbih}

The MIT-BIH Arrhythmia Database is a compilation of a wide range of common and less common but clinically significant arrhythmias, hence demanding to cope with different morphologies in each recording. The HDP concentration parameters were therefore initialised as $\gamma=1.0$ and $\alpha=0.1$. We set the LDS covariance priors $S_{\omega} = \varrho \Sigma_{\mathbf{y}_{d-1}}$ and $S_{\epsilon} = \Sigma_{\mathbf{y}_{d}}$, where we made use of the diagonalised data variance $\Sigma_{\mathbf{y}_{d}} = \frac{1}{D} \sum_{d=0}^{D} \mathbf{y}_{d}\mathbf{y}_{d}^{T}$ and the 1-step rolling variance of the data $\Sigma_{\mathbf{y}_{d-1}} = \frac{1}{D} \sum_{d=1}^{D} (\mathbf{y}_{d}-\mathbf{y}_{d-1})(\mathbf{y}_{d}-\mathbf{y}_{d-1})^{T}$. For the online approach, a small calibration segment at the beginning of the recording was used ($D=20$ beats), whereas the entire dataset was used for the offline approach. The factor $\varrho$ modulates the latent scaling matrix, biasing against the emergence of new clusters as $\varrho$ increases. As a rule of thumb, $\varrho=1.5$ can be advised for the online method, $\varrho=1.0$ for the offline method, and $\varrho=0.5$ for those recordings corresponding to patients with pacemakers (102, 104 and 217). Regarding the covariance function $k_{\theta}(t,t')$, we set the hyperparameters by maximising the marginal likelihood on the first beat of each cluster, from a common initial $\theta=(\sigma_{f}=300.0, \ell=1.0, \sigma_{n}=\sqrt{S_{\epsilon}})$, where $\sigma_{f}=300.0$ corresponds to the maximum signal deviation in a common ECG recording. We also set the hyperparameters of the covariance function $k_{\vartheta}(t,t')$ by maximising the marginal likelihood on the first beat of each cluster, from a common initial $\vartheta=(\sigma_{f}=1.0, \ell=4.0, \sigma_{n}=1.0)$.

\subsection{Runtime and memory analysis}
\label{app:runtime_analysis}

We also evaluated the empirical computational cost of HDP-GPC on the MIT-BIH Arrhythmia Database. The experiments were performed at record level under five configurations: online single-output without warping, offline single-output without warping, offline multi-output without warping, offline single-output with warping, and offline multi-output with warping. The latter corresponds to the most complete configuration considered in the ECG experiments. All experiments were executed on a 12th Generation Intel Core i5-12400F processor at 2.50 GHz with 32 GB of RAM.

\begin{figure}
    \centering
    \includegraphics[width=0.5\linewidth]{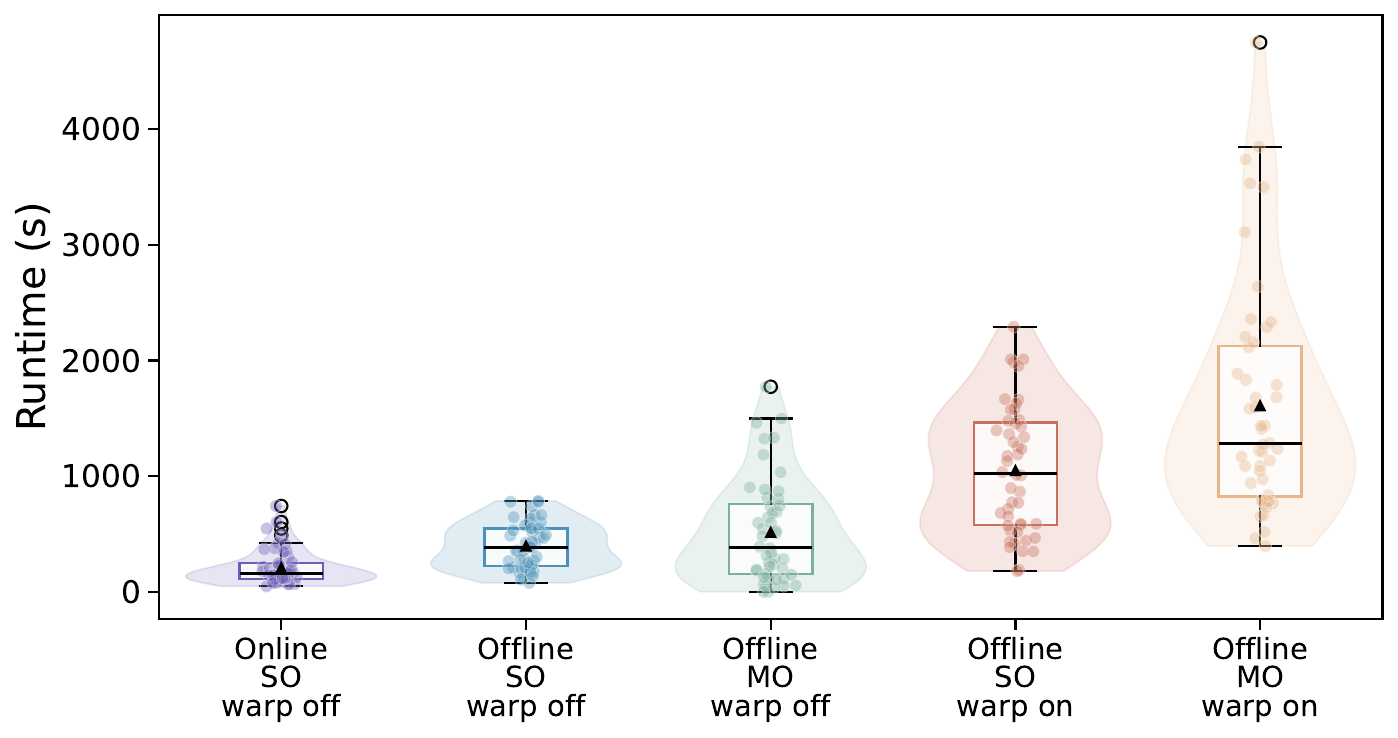}
    \caption{Distribution of execution times across MIT-BIH records for the five inference configurations. Violin shapes summarise empirical runtime density, boxplots show the interquartile range and median, triangles indicate the mean, and points correspond to individual records.}
    \label{fig:runtime_distribution_appendix}
    \Description[Runtime distributions by inference configuration]{
    Violin and box plots comparing execution times across five inference configurations. The online single-output configuration without warping has the lowest runtimes, while the offline multi-output configuration with warping has the largest runtimes and widest variability.
}
\end{figure}

Figure \ref{fig:runtime_distribution_appendix} summarises the distribution of execution times across records. The ordering is consistent with the structure of the inference algorithms. The online single-output no-warp variant is the fastest configuration, since it processes each new beat sequentially and does not revisit previous assignments. The offline no-warp variants are more expensive because they condition on the complete record, while the multi-output version adds a moderate overhead due to the evaluation and aggregation of lead-specific likelihoods. The largest increase is caused by warping: estimating a monotone alignment for each candidate morphology introduces an additional optimisation step, making the warped configurations substantially slower. The offline multi-output warped version has the largest median runtime and the largest variability, reflecting the combined cost of batch inference, multiple outputs, cluster comparison, and latent alignment.

\begin{figure}
    \centering
    \includegraphics[width=0.5\linewidth]{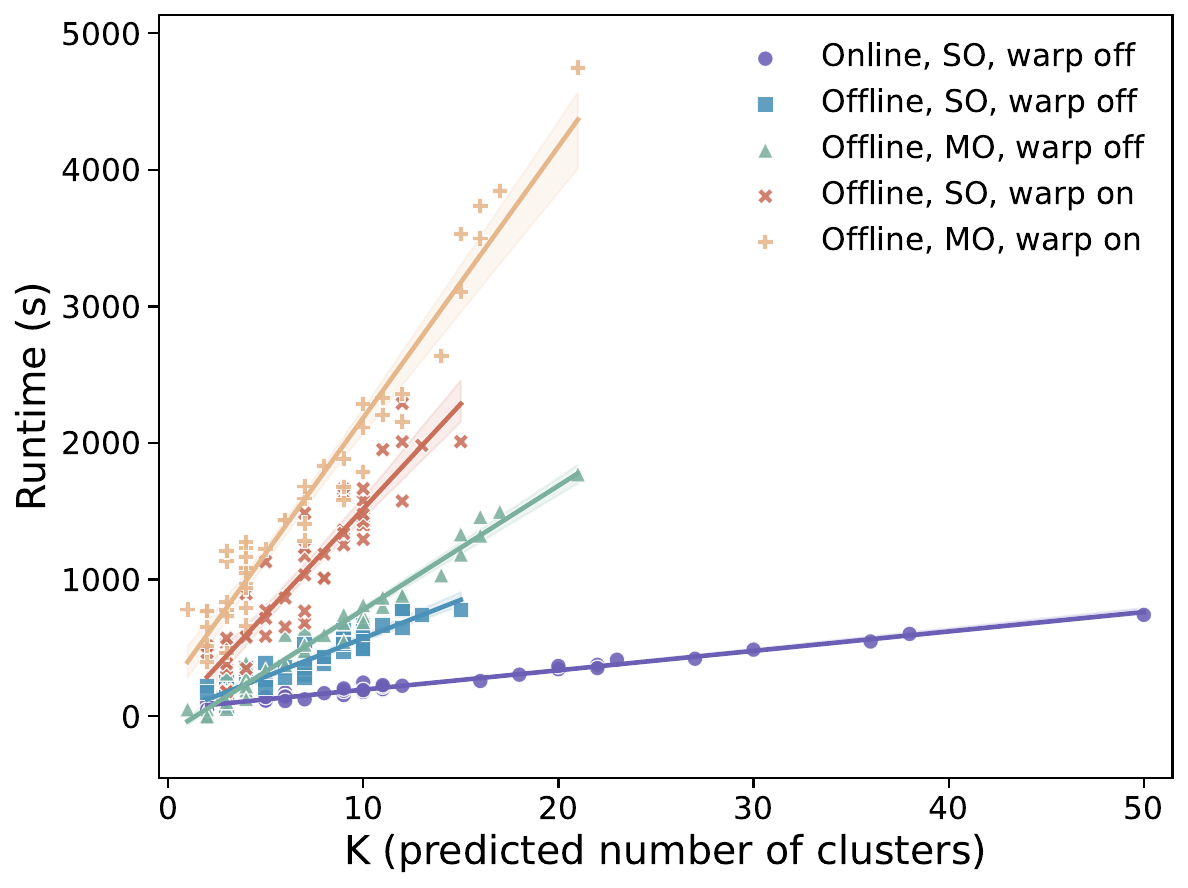}
    \caption{Empirical runtime scaling with the inferred number of clusters on the MIT-BIH Arrhythmia Database. Points correspond to individual records, solid lines show least-squares fits within each configuration, and shaded regions indicate bootstrap confidence envelopes.}
    \label{fig:runtime_vs_K_appendix}
    \Description[Runtime versus number of clusters]{
    Scatter plot of runtime against the inferred number of clusters for five inference configurations. Fitted trend lines increase with the number of clusters, and the warped and multi-output configurations occupy higher runtime ranges.
}
\end{figure}

\begin{figure}
    \centering
    \includegraphics[width=0.5\linewidth]{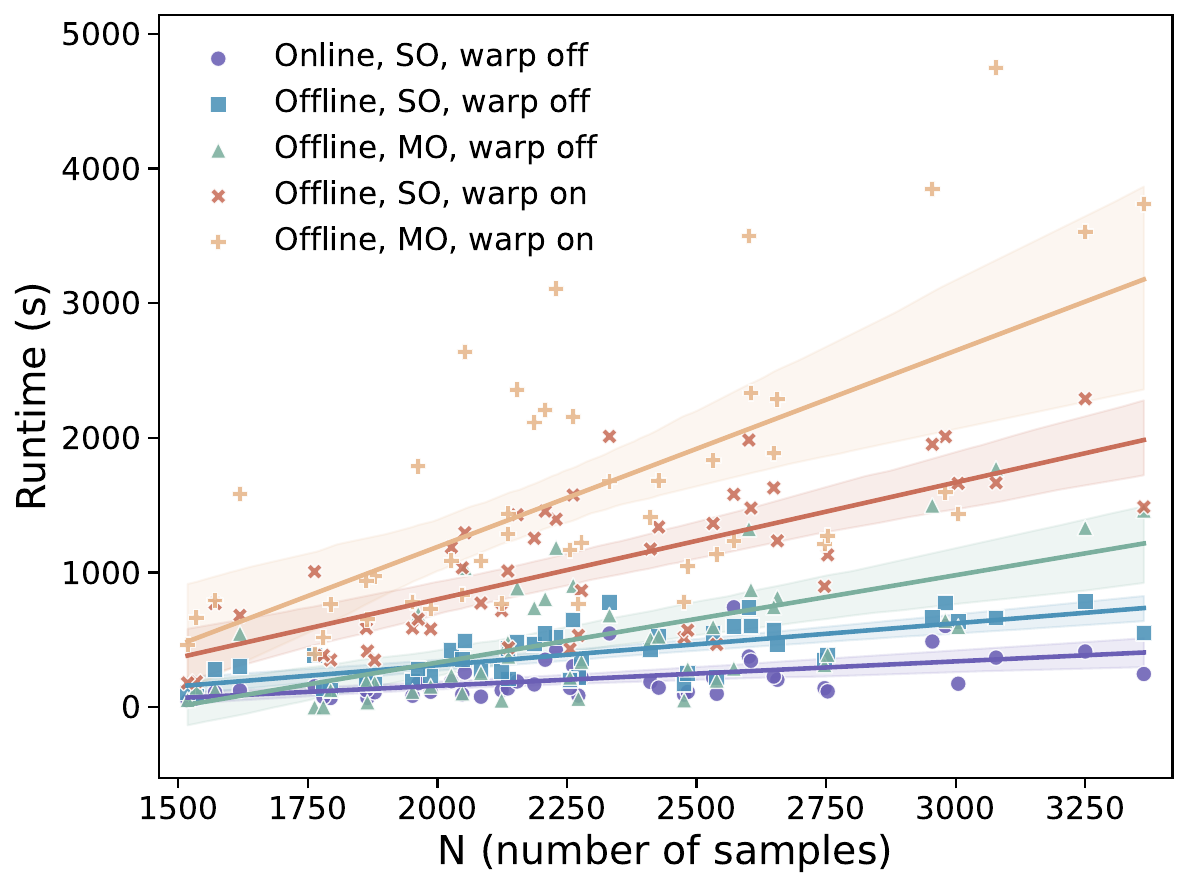}
    \caption{Empirical runtime scaling with the number of beats on the MIT-BIH Arrhythmia Database. Points correspond to individual records, solid lines show least-squares fits within each configuration, and shaded regions indicate bootstrap confidence envelopes.}
    \label{fig:runtime_vs_N_appendix}
    \Description[Runtime versus number of beats]{
    Scatter plot of runtime against the number of ECG beats for five inference configurations. Fitted trend lines show increasing computational cost for longer records, with offline warped configurations requiring the most time.
}
\end{figure}

Figure~\ref{fig:runtime_vs_K_appendix} shows the empirical dependence of runtime on the inferred number of clusters \(K\) and figure~\ref{fig:runtime_vs_N_appendix} on the number of beats \(N\). Runtime increases approximately linearly with both quantities, although the separation between configurations is more informative than the precise slope. The dependence on \(K\) is particularly relevant because each additional active morphology increases the number of candidate likelihood evaluations, dynamical updates, and, when enabled, warp estimations. The dependence on \(N\) is also positive, as longer records require more filtering, smoothing and assignment updates. In both plots, the online single-output no-warp configuration remains clearly separated from the offline variants, while the offline multi-output warped configuration occupies the upper envelope of the computational cost.

\begin{figure}
    \centering
    \includegraphics[width=0.75\textwidth]{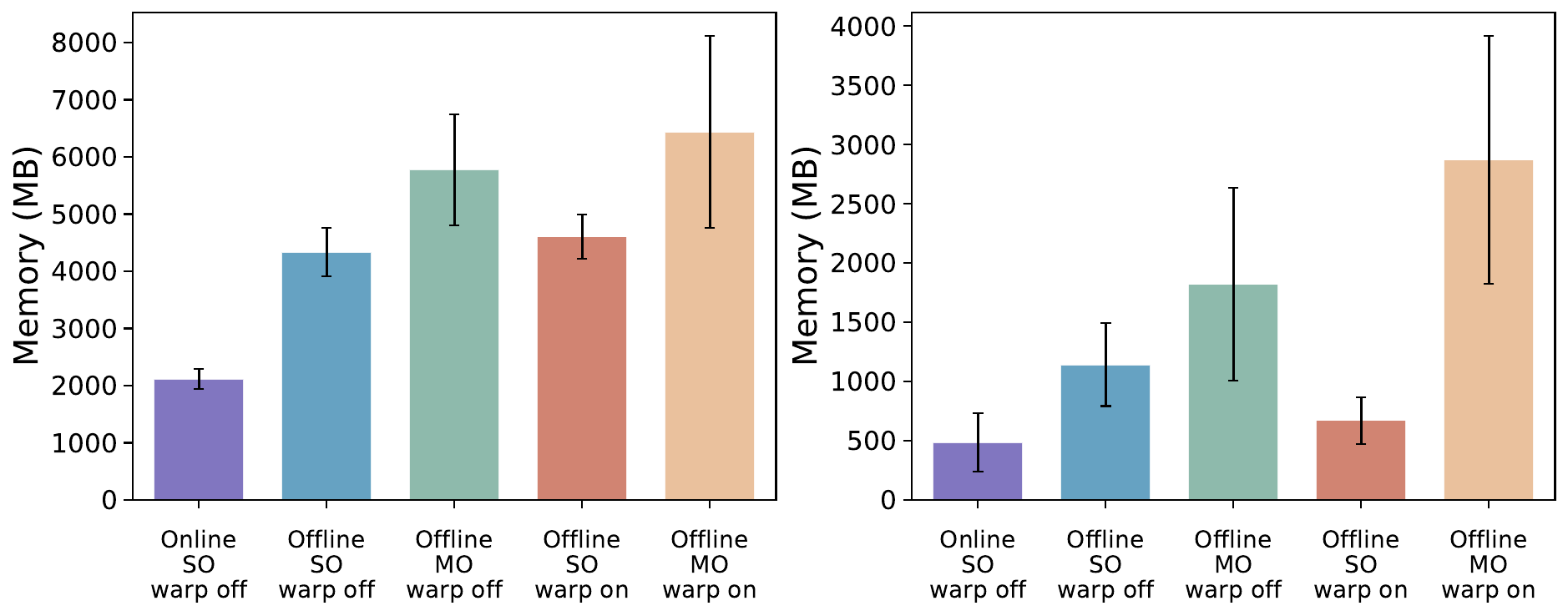}
    \caption{Memory usage by inference configuration. The left panel reports peak CPU resident set size, and the right panel reports the increase in resident memory during execution. Bars show means across records and error bars indicate one standard deviation.}
    \label{fig:memory_summary_appendix}
    \Description[Memory usage by inference configuration]{
    Pair of bar charts summarising memory usage across inference configurations. The left chart reports peak resident memory, and the right chart reports the increase in resident memory during execution. Offline, multi-output, and warped configurations show higher memory usage.
}
\end{figure}

The same hierarchy is observed in memory usage, as shown in Figure \ref{fig:memory_summary_appendix}. The online single-output no-warp configuration is the most memory efficient. offline inference requires a larger footprint because the complete sequence and posterior quantities must be retained during batch updates. Multi-output inference increases memory usage by storing and updating output-specific likelihood terms, and the warped configurations require additional structures associated with the alignment step. The offline multi-output warped configuration therefore has the highest peak resident memory and the largest memory increase during execution.

Overall, the runtime analysis shows the expected trade-off between modelling completeness and computational cost. The online single-output no-warp variant is the most appropriate configuration for large-scale or time-sensitive processing. The offline no-warp variants provide a stronger global clustering pass at moderate additional cost. The full offline multi-output warped model is the most expensive configuration, but it is also the most expressive one: it jointly exploits both ECG leads, estimates temporal alignment, and conditions cluster assignments on the complete record. Thus, it is best suited to offline analyses where interpretability, morphology quality and alignment are prioritised over execution time.

\section{Wave-spectra hyperparameters}
\label{app:wave_hyperparameters}

For the wave-spectra experiment, the same hyperparameter interpretation described in Section \ref{app:hyperparams_mitbih} was used. Since all spectra are defined on a common frequency support, no temporal warping was used. The model was therefore fitted with the multi-output offline HDP-GPC formulation, using each directional bin as one output channel and sharing the same discrete state sequence across outputs.

The GP emission prior used a squared-exponential covariance function. The initial output scale was set to $\sigma_f = 1.0$, with initial length-scale $\ell = 10^{-2}$. The GP hyperparameters were shared across the directional outputs. This choice enforces a common notion of smoothness over frequency while allowing each directional slice to contribute its own emission likelihood.

The LDS covariance scales were initialised from the data-driven estimators. We set the LDS covariance priors $S_{\omega} = \Sigma_{\mathbf{y}_{d-1}}$ and $S_{\epsilon} = \Sigma_{\mathbf{y}_{d}}$, where we made use of the diagonalised data variance described in Section \ref{sec:hyperparameters}, and the projection matrices were initialised as in Section \ref{sec:hyperparameters}, with $\mathbf{M}_A=\mathbf{I}$ and $\mathbf{M}_C=\mathbf{I}$. The HDP concentration parameters were not modified specifically for the wave-spectra experiment.

\section{Detailed Pedestrian Trajectory Clustering}
\label{app:pedestrian_clusters}

\begin{figure}
    \centering
    \includegraphics[width=0.65\textwidth]{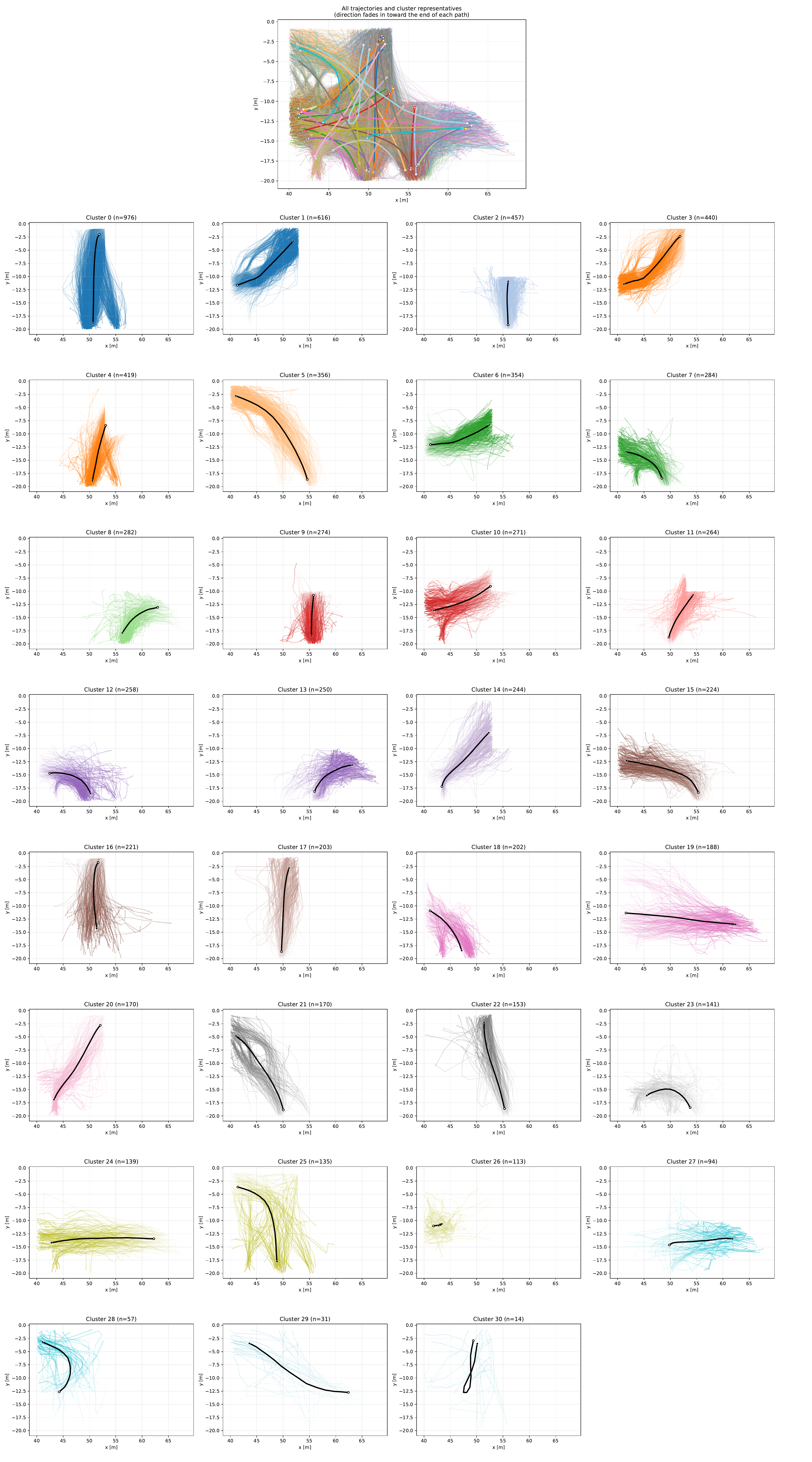}
    \caption{Complete pedestrian trajectory clustering obtained with the static HDP-GPC model. The top panel shows all trajectories and the inferred GP representatives. The remaining panels show the individual route families separately. The black curve denotes the GP representative of each cluster, and coloured curves denote the assigned trajectories. Direction is represented by fading each trajectory towards its endpoint; for the representatives, the open marker indicates the initial point and the filled marker indicates the final point.}
    \label{fig:pedestrian_clusters_appendix}
    \Description[Detailed pedestrian trajectory clusters]{
    Large grid of pedestrian trajectory clusters. The top panel shows all trajectories and their representatives together, while the remaining panels show individual route families. In each cluster panel, coloured curves represent assigned pedestrian paths and a black curve represents the Gaussian-process route representative.
}
\end{figure}

Figure \ref{fig:pedestrian_clusters_appendix} provides the complete cluster-level visualisation corresponding to the pedestrian trajectory experiment discussed in Section \ref{sec:applications}. The main text reports a compact overview for readability, whereas this appendix shows the individual route families inferred by the static HDP-GPC model. The top panel overlays all trajectories and cluster representatives, and the remaining panels display each cluster separately. The black curve denotes the GP representative of the route family, while the coloured trajectories correspond to the paths assigned to that cluster.

The detailed panels show that the inferred clusters correspond to interpretable movement patterns in the station layout. Large clusters capture dominant corridors, while smaller clusters represent less frequent but coherent route families, such as curved movements, side accesses, or short local displacements. This supports the interpretation in the main text: after collapsing the Markovian transition structure, the static HDP-GPC model behaves as an exchangeable nonparametric mixture of multi-output GP trajectories, discovering route families without requiring a predefined number of clusters or hand-crafted origin--destination labels.

\subsection{Pedestrian-trajectory hyperparameters}
\label{app:pedestrian_hyperparameters}

For the pedestrian-trajectory experiment, the static reduction of HDP-GPC was used. Therefore, the transition rows were collapsed to the global HDP weights and the dynamical matrices were fixed to the identity. Consequently, the LDS parameters do not encode temporal evolution between consecutive trajectories; instead, each route family is represented directly by its GP emission model.

The GP emission prior used a squared-exponential covariance function on the normalised trajectory-progress support. The initial length-scale was set to $\ell = 1.0$. The initial output scale was computed from the maximum absolute value of the resampled trajectory data $\sigma_f = \max_{n} |\mathbf{y}_{n}|$, where $n$ indexes trajectories. GP hyperparameters were shared across the two spatial outputs.

The LDS hyperparameters were initialised from empirical moment estimators, following the same criterion used in Section \ref{app:hyperparams_mitbih}: $S_{\omega} = \Sigma_{\mathbf{y}_{d-1}}$ and $S_{\epsilon} = \Sigma_{\mathbf{y}_{d}}$. Temporal warping was disabled, since all trajectories were already resampled onto a common progress variable $\tau\in[0,1]$.

\end{document}